\documentclass[10pt,journal,compsoc]{IEEEtran}
\ifCLASSOPTIONcompsoc
  \usepackage[nocompress]{cite}
\else
  \usepackage{cite}
\fi
\ifCLASSINFOpdf
\else
\fi
\usepackage{amsmath}
\usepackage{amsfonts}
\usepackage{tabularx}
\usepackage{diagbox}
\newcommand\textsharp{\raisebox{\depth}{\#}}

\makeatletter
\newcommand{\algmultiline}[1]{%
  \begin{tabularx}{\dimexpr\linewidth-\ALG@thistlm}[t]{@{}X@{}}
    #1
  \end{tabularx}
}
\makeatother

\newcommand{\makecomments}[1]{\ignorespaces}

\usepackage{algpseudocode}
\usepackage{algorithm}
\usepackage{algorithmicx}
\usepackage{bm}

\renewcommand{\mathbf}{\boldsymbol}

\usepackage{array}
\newcolumntype{L}[1]{>{\raggedright\let\newline\\\arraybackslash\hspace{0pt}}m{#1}} 
\newcolumntype{C}[1]{>{\centering\let\newline\\\arraybackslash\hspace{0pt}}m{#1}}
\newcolumntype{R}[1]{>{\raggedleft\let\newline\\\arraybackslash\hspace{0pt}}m{#1}}

\usepackage{graphicx}
\usepackage{subfigure}
\usepackage{xcolor}
\usepackage{url}

\usepackage{amsmath,amssymb,amsthm}
\usepackage{multirow}
\newtheorem{theorem}{Theorem}
\newtheorem{definition}[theorem]{Definition}
\newtheorem{lemma}[theorem]{Lemma}
\newtheorem{corollary}[theorem]{Corollary}

\begin{document}
%
\title{Learning and Meshing from Deep \\ Implicit Surface Networks Using an Efficient Implementation of Analytic Marching}
%
%
%
%

\author{Jiabao~Lei,~Kui~Jia,~and~Yi~Ma,~\IEEEmembership{Fellow,~IEEE}
  \IEEEcompsocitemizethanks{
  \IEEEcompsocthanksitem J. Lei and K. Jia are with the School of Electronic and Information Engineering, South China University of Technology, Guangzhou, China. E-mails: eejblei@mail.scut.edu.cn, kuijia@scut.edu.cn. 
  \IEEEcompsocthanksitem Y. Ma is with the Department of Electrical Engineering and Computer Sciences, University of California at Berkeley, Berkeley, CA 94720-1770 USA. E-mails: yima@eecs.berkeley.edu.
  \IEEEcompsocthanksitem Correspondence to: K. Jia.  \protect
}}

\IEEEtitleabstractindextext{%
  \begin{abstract}
    Reconstruction of object or scene surfaces has tremendous applications in computer vision, computer graphics, and robotics.
    The topic attracts increased attention with the emerging pipeline of deep learning surface reconstruction,
    where implicit field functions constructed from deep networks (e.g., multi-layer perceptrons or MLPs) are proposed for generative shape modeling. In this paper, we study a fundamental problem in this context about recovering a surface mesh from an implicit field function whose zero-level set captures the underlying surface. To achieve the goal, existing methods rely on traditional meshing algorithms (e.g., the de-facto standard marching cubes); while promising, they suffer from loss of precision learned in the implicit surface networks, due to the use of discrete space sampling in marching cubes.
    Given that an MLP with activations of Rectified Linear Unit (ReLU) partitions its input space into a number of linear regions,
    we are motivated to connect this local linearity with a same property owned by the desired result of polygon mesh.
    More specifically, we identify from the linear regions, partitioned by an MLP based implicit function, the \emph{analytic cells} and \emph{analytic faces}
    that are associated with the function's zero-level isosurface. We prove that under mild conditions,
    the identified analytic faces are guaranteed to connect and form a \emph{closed, piecewise planar surface}.
    Based on the theorem, we propose an algorithm of \emph{analytic marching}, which marches among analytic cells to \emph{exactly} recover the mesh captured by an implicit surface network.
    We also show that our theory and algorithm are equally applicable to advanced MLPs with shortcut connections and max pooling.
    Given the parallel nature of analytic marching, we contribute \texttt{AnalyticMesh}, a software package that supports efficient meshing of implicit surface networks via CUDA parallel computing, and mesh simplification for efficient downstream processing. 
    We apply our method to different settings of generative shape modeling using implicit surface networks. Extensive experiments demonstrate our advantages over existing methods in terms of both meshing accuracy and efficiency. Codes are at \url{https://github.com/Karbo123/AnalyticMesh}.
  \end{abstract}

  }

\maketitle

\IEEEdisplaynontitleabstractindextext

%
\IEEEpeerreviewmaketitle




\section{Introduction}\label{SecIntro}

\IEEEPARstart{C}{reation} of 3D content prepares geometric data useful for analysis and processing in many scientific fields.
For example, in computer vision and robotics, object or scene surface reconstruction via simultaneous localization and mapping
\cite{Newcombe2011} enables robotic manipulation, indoor navigation, and urban modeling; in computer graphics,
reconstruction of continuous surface from discrete raw scanning is the first step in computer-aided design, virtual/augmented reality,
and movie production. The geometric data created in these applications are of 2-dimensional manifold embedded in the 3D space.
In this work, we are particularly interested in those data of closed manifolds representing, e.g., the boundary of a 3D solid.

As a mathematical notion of geometry, a continuous surface manifold is difficult to be modeled directly; in practice, it is approximated as different representations,
such as spline surface, subdivision surface, or polygon mesh \cite{Botsch2010}. Among them, the polygon mesh is arguably the most popular representation proposed
in the literature, which approximates a smooth surface explicitly as a piecewise, linear function; for example, the most typical triangle mesh is defined as a
collection of connected faces, each of which has three vertices that uniquely determine plane parameters of the face in the 3D space. Given the parametric mappings
specified by planar faces of a polygon mesh, the representation is advantageous in surface evaluation and rendering; however, it is usually difficult to obtain a mesh directly,
especially for topologically complex surface; queries of points inside or outside the surface are expensive as well.
As an alternative, one may resort to implicit surface representations, such as signed distance function (SDF) \cite{Curless1996,Park2019} or
occupancy field (OF) \cite{Chen2018,Mescheder2019}, which subsume a surface as the zero-level isosurface in the function field; other implicit representations
include discrete volumes and those based on algebraic \cite{Blinn1982,Nishimura1985,Wyvill1986} and radial basis functions \cite{Carr2001,Carr1997,Turk1999}.
To obtain a surface mesh, the continuous field is often sampled discretely as a regular grid of voxels, followed by the de-facto standard algorithm of
marching cubes \cite{Lorensen1987}. Efficiency and precision of marching cubes can be improved on a hierarchically sampled structure of octree via algorithms
such as dual contouring \cite{Ju2002}.

Implicit functions are traditionally implemented based on moving least squares \cite{Kolluri08}. More recently, methods of deep learning surface reconstruction \cite{Park2019,Chen2018,Mescheder2019} propose to leverage the great modeling capacities of deep networks (e.g., Multi-Layer Perceptrons (MLPs) based on Rectified Linear Unit (ReLU) \cite{Glorot2011}) to learn implicit fields.
Given a learned field, they again take a final step of marching cubes to obtain the mesh result.
While promising, the final step of marching cubes recovers a mesh that is only an approximation of the surface captured by the learned implicit network;
more specifically, it suffers from a trade-off of sampling efficiency and recovery precision, due to the discretization nature of the marching cubes algorithm.
The very recent deep models using soft ReLU \cite{Gropp2020} or sine/cosine activation functions \cite{Sitzmann2019} suffer from this limitation as well.

To address the limitation, we are motivated from the established knowledge that a ReLU based MLP partitions its input space into a number of linear regions \cite{Montufar2014}; this connects with the locally linear property of polygon mesh. Given an MLP based implicit function, we identify from its partitioned linear regions the \emph{analytic cells} and \emph{analytic faces} that are associated with the function's zero-level isosurface. Assuming that such an implicit function learns its zero-level isosurface as a \emph{closed, piecewise planar surface}, we characterize theoretical conditions under which analytic faces of the implicit function \emph{connect and exactly form} the surface mesh. Based on our theorem, we propose an algorithm of \emph{analytic marching}, which marches among analytic cells to recover the \emph{exact mesh} of the closed, piecewise planar surface captured by a learned MLP. Our choices of MLPs also include those with shortcut connections and max pooling. The proposed analytic marching algorithm can be naturally implemented in parallel, for which we contribute \texttt{AnalyticMesh}, a software package that supports efficient meshing of implicit surface networks via CUDA parallel computing, and postprocessing of mesh simplification. We apply our meshing algorithm in the contexts of either direct shape decoding of raw point observations, or learning to reconstruct novel shape instances using global or local decoders. Experiments on benchmark 3D object repositories show the advantages of our meshing algorithm over existing ones.

\subsection{Relations with the Literature}

The problem studied in this work is closely related to the following three lines of research.

\vspace{0.1cm}
\noindent\textbf{Implicit Surface Representations.}
An implicit surface representation is defined as the zero-level set of an scalar-valued implicit function. Earlier methods take a divide-and-conquer strategy that represents the surface using atom functions. For example, blobby molecule \cite{Blinn1982} is proposed to approximate each atom by a gaussian potential, and a piecewise quadratic meta-ball \cite{Nishimura1985} is used to approximate the gaussian, which is improved via a soft object model in \cite{Wyvill1986} by using a sixth degree polynomial. Radial basis function (RBF) is an alternative to the above algebraic functions. RBF-based approaches \cite{Carr2001,Carr1997,Turk1999} place the function centers near the surface and are able to reconstruct a surface from a discrete point cloud. 
It has been recently discovered that deep networks, owing to their great modeling capacities, are able to learn implicit surface fields very effectively. DeepSDF \cite{Park2019} trains ReLU based MLPs as signed distance functions. IMNet \cite{Chen2018} and OccNet \cite{Mescheder2019} learn similar types of networks as occupancy fields. Deep implicit surface networks are also used in \cite{Xu2019} for surface reconstruction from as few as a single image. Other than ReLU based networks, smooth activations such as soft ReLU \cite{Gropp2020} or sine/consine functions \cite{Sitzmann2019} have been showing the new promise for learning smoother surfaces via implicit fields. We focus on ReLU based networks in the present work.

\vspace{0.1cm}
\noindent\textbf{Mesh Conversions from Implicit Fields.}
The conversion from an implicit representation to an explicit surface mesh is called isosurface extraction. Probably the simplest approach is to directly convert an implicit volume via greedy meshing (GM). The de-facto standard algorithm of marching cubes (MC) \cite{Lorensen1987} builds from the implicit function a discrete volume around the surface of interest, and then computes mesh vertices on the edges of the volume; due to its discretization nature, mesh results of the algorithm are often short of sharp surface details. Algorithms similar to MC include marching tetrahedra (MT) \cite{Doi1991} and dual contouring (DC) \cite{Ju2002}. MT divides a voxel into six tetrahedrons and calculates the vertices on edges of each tetrahedron; DC utilizes gradients to estimate positions of vertices in a cell and extracts meshes from adaptive octrees. All these methods suffer from a trade-off of precision and efficiency, due to their necessity to sample discrete points from the 3D space.

\vspace{0.1cm}
\noindent\textbf{Local Linearity of MLPs.}
Among works studying representational complexities of deep networks, {Mont{\'u}far} et al. \cite{Montufar2014} and Pascanu et al. \cite{Pascanu2013} investigate how a ReLU or maxout based MLP partitions its input space into a number of linear regions, and bound this number via quantities relevant to network depth and width. The region-wise linear mapping is explicitly established in \cite{Jia2019} in order to analyze generalization properties of deep networks. A closed-form solution termed OpenBox is proposed in \cite{Chu2018} that computes consistent and exact interpretations for piecewise linear deep networks. The present work leverages the locally linear properties of ReLU based MLPs and studies how the zero-level isosurface can be extracted from such an MLP based implicit function.

\subsection{Contributions}

A preliminary version of this work appears in \cite{Lei2020},  where for the first time, we establish the analytic relations between an MLP based implicit function and its captured zero-level isosurface; we present in \cite{Lei2020} a theorem that guarantees exact meshing from deep implicit surface networks, and a corresponding meshing algorithm. We re-state its technical contributions as follows.
\begin{enumerate}
  \item Given that an MLP with ReLU activation partitions its input space into a number of linear regions, we identify from these regions \emph{analytic cells} and \emph{analytic faces} that are associated with zero-level isosurface of an implicit function constructed from such an MLP; we characterize the theoretical conditions under which the identified analytic faces are guaranteed to connect and form a \emph{closed, piecewise planar surface}.
  \item Based on the above analytic meshing theorem, we propose an algorithm of \emph{analytic marching}, which marches among analytic cells to \emph{exactly} recover the mesh captured by an implicit surface network. We empirically verify that the proposed meshing algorithm achieves a precision upper-bounding those achieved by existing algorithms.
\end{enumerate}
In the present paper, we extend the theoretical analysis in \cite{Lei2020} for more advanced MLP architectures, and contribute techniques to improve the efficiency of analytic marching. These extensions enable us to apply our proposed method to learning and meshing novel and complex shape instances. In addition, we augment the paper presentation with motivation of theory and intuitive illustrations. We finally summarize our new contributions as follows.
\begin{enumerate}
  \item We present analyses that make the analytic meshing theorem in \cite{Lei2020} applicable to more advanced MLP architectures, including those with shortcut connections and max pooling operations. These extensions support a richer set of architectural designs for learning and exactly meshing complex surface shapes with analytic marching.

  \item We contribute techniques to improve the efficiency of analytic marching, including \emph{parallel marching} with CUDA implementation, efficient initialization schemes respectively customized for signed distance field and occupancy field, and mesh simplification for efficient downstream processing. Implementations of these techniques are included in \texttt{AnalyticMesh}, a software package accessible at \url{https://github.com/Karbo123/AnalyticMesh}.

  \item We apply our method to different contexts of generative shape modeling using implicit surface networks; we consider both direct shape decoding of raw point observations, and learning to reconstruct novel shape instances using global or local shape decoders. Extensive experiments demonstrate the advantages of our meshing algorithm over existing ones in terms of both accuracy and efficiency.
\end{enumerate}


\section{Problem Statement and Motivation}
\label{SecProbAndMotivation}

This paper studies the fundamental problem of recovering an \emph{explicit} representation $\mathcal{Z}$ of an underlying surface $\mathcal{M}$ from some, possibly learned, \emph{implicit} surface function. We focus our surface of interest on those representing the boundary of a non-degenerate 3D solid whose nature is a continuous and closed 2-dimensional manifold embedded in the Euclidean space $\mathbb{R}^3$ ; such a solid has no infinitely thin parts and its boundary surface properly separates the interior and exterior of the solid (cf. Fig. 1.1 in \cite{Botsch2010} for an illustration). Among choices of explicit surface representation, a polygon mesh is the most popular one defined as $\mathcal{Z} = \{\mathcal{V}, \{\mathcal{P}\} \}$, where $\mathcal{V} = \{ \bm{v} \in \mathbb{R}^3 \}$ contains the mesh vertices and $\{\mathcal{P} \subset \mathbb{P}^2 \}$ denotes the collection of connected polygon faces, each of which contains a coplanar set of vertices \footnote{For simplicity, we omit edges in the definition of polygon mesh. Edges can be inferred as boundaries of polygon faces. In this work, we consider the non-degenerate case that any edge has no more than two incident faces and any vertex is incident to no more than one fan of faces. }. Any planar face thus defines an explicit mapping $\bm{g}_{\mathcal{P}}: \Omega \rightarrow \mathbb{R}^3$ from the domain $\Omega$ (e.g., $\Omega \subset \mathbb{R}^2$) to a plane $\mathbb{P}^2 \subset \mathbb{R}^3$; consequently, the mesh $\mathcal{Z}$ becomes a piecewise linear approximation of the underlying $\mathcal{M}$. Recent results \cite{Groueix2018,Wang2018,Tang2019,Pan2019} show that the collection $\{ \bm{g}_{\mathcal{P}} \}$ of explicit mapping functions can be effectively learned as one or several deep networks, which are trained to generate a surface mesh via vertex deformation. However, topologies of the resulting meshes are restricted by those defined on the input domain $\Omega$; queries of points inside or outside the surface are expensive as well.

As an alternative, one may resort to implicit surface representations, such as signed distance function (SDF) \cite{Park2019} or occupancy field \cite{Mescheder2019,Chen2018}. Implicit representations enjoy the benefits of modeling smooth and topologically complex surfaces. Let $F: \mathbb{R}^3 \rightarrow \mathbb{R}$ denote a scalar-valued, implicit field function, and a surface is formally defined as its zero-level isosurface $\{ \bm{x} \in \mathbb{R}^3 | F(\bm{x}) = 0 \}$.
\footnote{
  Given an implicit field function $F: \mathbb{R}^3 \rightarrow \mathbb{R}$, the surface of interest is more precisely defined as
  \begin{equation}\label{EqnPreciseImFieldDefinition}
    \left\{ \boldsymbol{x} \in \mathbb{R}^3 \mid F(\boldsymbol{x}) = 0,
    \left| \nabla_{\bm{x}} F(\bm{x}) \right| \neq \bm{0} \right\},
  \end{equation}
  where the constraint $\left| \nabla_{\bm{x}} F(\bm{x}) \right| \neq \bm{0}$ ensures that (\ref{EqnPreciseImFieldDefinition}) indeed defines a zero-crossing isosurface. When $F$ implements a signed distance function, $|F(\bm{x})|$ measures the distance of any $\bm{x} \in \mathbb{R}^3$ to the surface, and by convention we have $F(\mathbf{x}) < 0$ for points inside the surface and $F(\mathbf{x}) > 0$ for those outside. When $F$ represents an occupancy field, it implements a mapping $\mathbb{R}^3 \rightarrow \{0, 1\}$, which assigns each $\bm{x} \in \mathbb{R}^3$ a binary occupancy value indicating the exterior ($F(\bm{x}) = 0$) or interior ($F(\bm{x}) = 1$) status of $\bm{x}$.
}
While $F$ can be realized using radial basis functions \cite{Carr2001,Carr1997,Turk1999} or be approximated as a regular grid of voxels (i.e., a volume), in this work, we are particularly interested in implementing $F$ using deep networks, e.g., Multi-Layer Perceptrons (MLPs) with Rectified Linear Units \cite{Glorot2011}, which become an increasingly popular choice in recent works of deep learning surface reconstruction \cite{Park2019,Chen2018,Xu2019}. These methods achieve the state-of-the-art performance in terms of surface modeling, and they typically take a final step of marching cubes \cite{Lorensen1987} to recover the surface meshes, by sampling and evaluating a regular grid of discrete points in the 3D space. As stated in Section \ref{SecIntro}, the final step of marching cubes recovers a mesh that is only an approximation of the surface captured by $F$; more specifically, it suffers from a trade-off of sampling efficiency and recovery precision, due to the discretization nature of the marching cubes algorithm.

\begin{figure*}[!htbp]
  \vskip 0.1in
  \begin{center}
    \includegraphics[scale=0.16]{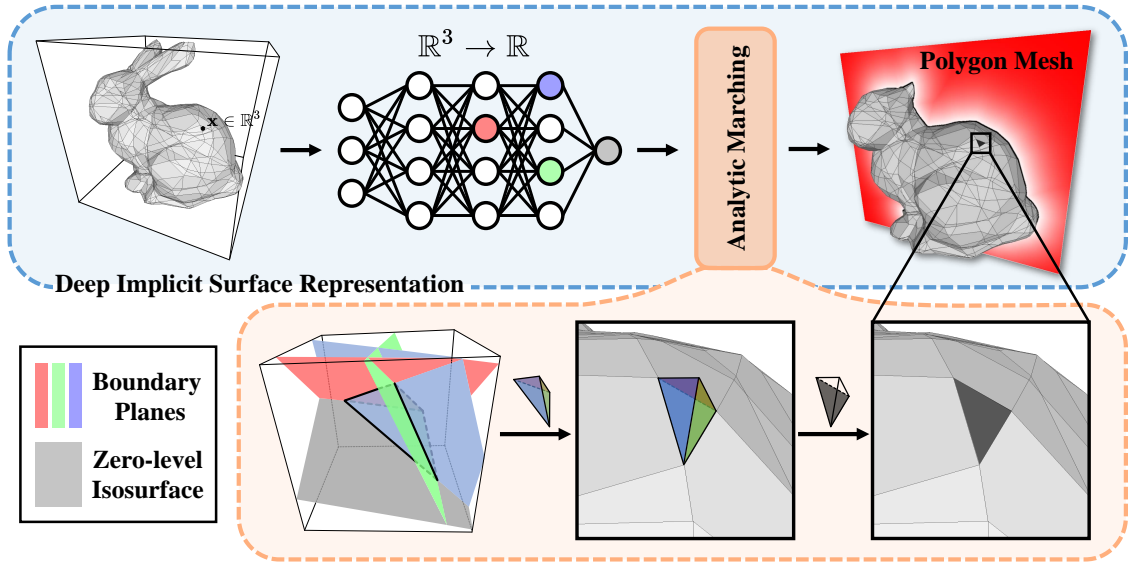} \vspace{-0.4cm}
    \caption{An illustration on the intrinsic connection between a polygon mesh and the local linearities of its capturing deep implicit surface network. The polygon mesh is formed by the intersection between the zero-level isosurface of the implicit surface network and some of its partitioned linear regions in the input Euclidean space. }
    \label{FigIntuition}
  \end{center}
  \vskip -0.1in
\end{figure*}

In this work, we aim to address this limitation by developing a meshing algorithm whose nature is completely different from the discrete meshing family of marching cubes \cite{Lorensen1987,Doi1991,Ju2002}. As stated above, most of existing deep implicit surface functions are based on MLPs. We note that a ReLU based MLP partitions its input space into a number of linear regions \cite{Montufar2014}; consequently, the zero-level isosurface $\{ \bm{x} \in \mathbb{R}^3 | F(\bm{x}) = 0 \}$ of such a function $F$ is embedded in the input space $\mathbb{R}^3$ and is intersected by (some of) the partitioned linear regions. Given the \emph{piecewise planar/linear $\{ \bm{g}_{\mathcal{P}} \}$ of a polygon mesh $\mathcal{Z}$} and the \emph{locally linear mappings defined by an MLP based $F$}, we are motivated to connect the local linearities of the two worlds and \emph{analytically} identify the linear regions intersected by the zero-level isosurface; we expect these intersections to form $\mathcal{Z}$ that is an \emph{exact} meshing solution from $F$. In the present section, we give formal statement of the problem and our motivation. Fig. \ref{FigIntuition} illustrates the intuition.

\subsection{Polygon Mesh as a Piecewise Linear Surface Representation}
\label{SecPolygonMesh}

Assume we have a polygon mesh $\mathcal{Z} = \{\mathcal{V}, \{ \mathcal{P}\} \}$ embedded in the space $\mathbb{R}^3$. For any planar face $\mathcal{P}$, let $\{ \bm{v}_i \in \mathcal{V} | i = 1, \dots, n_{\mathcal{P}} \}$ be its defining vertices.
Given any three $\{ \bm{v}_i, \bm{v}_j, \bm{v}_k \}$ of these vertices, the plane on which the polygon segment $\mathcal{P}$ resides can be written as
\begin{equation}
  \bm{n}_{\mathcal{P}}^{\top} (\bm{x} - \bm{v}_i) = 0 \ \ \textrm{s.t.} \ \ \bm{n}_{\mathcal{P}} =
  \frac{ (\bm{V}\bm{V}^{\top})^{-1} \bm{V} \cdot \bm{1}_3 }{\Vert (\bm{V}\bm{V}^{\top})^{-1} \bm{V} \cdot \bm{1}_3 \Vert_{2}} ,
\end{equation}
where the matrix $\bm{V} = [\bm{v}_i, \bm{v}_j, \bm{v}_k]$ collects coordinates of the three vertices, $\bm{1}_3$ is a 3-dimensional vector with all its entries as the value of 1, $\bm{n}_{\mathcal{P}} \in \mathbb{R}^3$ is the plane kernel or normal vector, and $\bm{x} \in \mathbb{R}^3$ is any space point on the plane. The collection $\{ \bm{n}_{\mathcal{P}} \}$ thus gives a piecewise linear parameterization of $\mathcal{Z}$. We will show in the following that any ReLU based MLP $F$ has its zero-level isosurface as polygon segments embedded in $\mathbb{R}^3$, which motivates a possible solution of analytic meshing from $F$.

\subsection{The Local Linearity of Multi-Layer Perceptrons}
\label{SecLocalLinearityofMLP}

We first discuss how a ReLU based MLP, as a nonlinear function, partitions its input space into linear regions via compositional structure. The discussion is put in a general form by assuming an MLP of $L$ hidden layers that takes an input $\mathbf{x} \in \mathbb{R}^{n_0}$ from the space $\mathcal{X}$ and layer-wisely computes $\mathbf{x}_l = \mathbf{g} (\mathbf{W}_l \mathbf{x}_{l-1})$, where $l \in \{1, \dots, L\}$ indexes the layer, $\mathbf{x}_l \in \mathbb{R}^{n_l}$, $\mathbf{x}_0 = \mathbf{x}$, $\mathbf{W}_l \in \mathbb{R}^{n_l\times n_{l-1}}$, $\mathbf{g}$ is the point-wise ReLU activation, and we omit the network biases for notational simplicity. We also denote the intermediate feature space $\bm{g}(\mathbf{W}_l\mathbf{x}_{l-1})$ as $\mathcal{X}_l$ and $\mathcal{X}_0 = \mathcal{X}$. In the context of present paper, we have $\mathcal{X}_0 \subset \mathbb{R}^3$ and $n_0 = 3$.

The thus defined MLP can be compactly written as
\begin{equation}\label{EqnMLP}
  \mathbf{T}\mathbf{x} = \mathbf{g}(\mathbf{W}_{L}\ldots \mathbf{g}(\mathbf{W}_1\mathbf{x})) .
\end{equation}
Any $k^{th}$ neuron, $k \in \{1,\ldots, n_l\}$, of an $l^{th}$ layer of the MLP $\mathbf{T}$ specifies a pre-activation functional defined as
\begin{displaymath}
  a_{lk}(\mathbf{x}) = \pi_k \mathbf{W}_l\mathbf{g}(\mathbf{W}_{l-1}\ldots \mathbf{g}(\mathbf{W}_1\mathbf{x})) ,
\end{displaymath}
where $\pi_k$ denotes an operator that projects onto the $k^{th}$ coordinate. All the neurons at layer $l$ define a functional
\begin{displaymath}
  \mathbf{a}_{l}(\mathbf{x}) =  \mathbf{W}_l\mathbf{g}(\mathbf{W}_{l-1}\ldots \mathbf{g}(\mathbf{W}_1\mathbf{x})) .
\end{displaymath}
We define the support of $\mathbf{T}$ as
\begin{equation}\label{EqnMLPSupport}
  \textrm{supp}(\mathbf{T}) = \{ \mathbf{x} \in \mathcal{X} | \mathbf{T}\mathbf{x} \not= \mathbf{0}\} ,
\end{equation}
which are instances of practical interest in the input space \footnote{Any instance $\mathbf{x} \in \mathcal{X}$ nullified by an MLP $\bm{T}$ of $L$ hidden layers defined as (\ref{EqnMLP}) would be less useful for downstream tasks, e.g., an implicit function constructed from $\bm{T}$.}.

For an intermediate feature space $\mathcal{X}_{l-1} \in \mathbb{R}^{n_{l-1}}$, each hidden neuron of layer $l$ specifies a hyperplane $\bm{H}$ that partitions $\mathcal{X}_{l-1}$ into two halves, and the collection of hyperplanes $\{\bm{H}_i\}_{i=1}^{n_l}$ specified by all the $n_l$ neurons of layer $l$ form a \emph{hyperplane arrangement} \cite{Peter1992}. These hyperplanes partition the space $\mathcal{X}_{l-1}$ into multiple linear regions whose formal definition is as follows.
\begin{definition}[Region/Cell]
  Let $\mathcal{A}$ be an arrangement of hyperplanes in $\mathbb{R}^m$. A region of the arrangement is a connected component of the complement $\mathbb{R}^m - \bigcup\limits_{\bm{H} \in \mathcal{A}}\bm{H}$. A region is a cell when it is bounded. 
\end{definition}
\noindent Classical result from \cite{Zaslavsky1975,Pascanu2013} shows that the arrangement of $n_l$ hyperplanes  gives at most $\sum_{j=0}^{n_{l-1}}{{n_l}\choose{j}}$ regions in $\mathbb{R}^{n_{l-1}}$. Given fixed $\{\mathbf{W}_l\}_{l=1}^{L}$, the MLP $\mathbf{T}$ partitions the input space $\mathcal{X} \in \mathbb{R}^{n_0}$ by its layers' recursive partitioning of intermediate feature spaces, which can be intuitively understood as a successive process of space folding \cite{Montufar2014}.

Let $\mathcal{R}(\mathbf{T})$, shortened as $\mathcal{R}$, denote the set of all linear regions/cells in $\mathbb{R}^{n_0}$ that are possibly achieved by $\mathbf{T}$. To have a concept on the maximal size of $\mathcal{R}$, we introduce the following functionals about activation states of neuron, layer, and the whole MLP.
\begin{definition}[State of Neuron/MLP]\label{DefinitionNeuronMLPState}
  For a $k^{th}$ neuron of an $l^{th}$ layer of an MLP $\mathbf{T}$, with $k \in \{1, \dots, n_l\}$ and $l \in \{1, \dots, L\}$, its state functional of neuron activation is defined as
  \begin{equation}\label{EqnNeuronStateFunctional}
    s_{lk}(\mathbf{x}) =
    \begin{cases}
      1 & \text{if $a_{lk}(\mathbf{x}) > 0$}      \\
      0 & \text{if $a_{lk}(\mathbf{x}) \leq 0$} ,
    \end{cases}
  \end{equation}
  which gives the state functional of layer $l$ as
  \begin{equation}\label{EqnLayerStateFunctional}
    \mathbf{s}_l(\mathbf{x}) = [s_{l1}(\mathbf{x}), \dots, s_{ln_l}(\mathbf{x})]^{\top} ,
  \end{equation}
  and the state functional of MLP $\mathbf{T}$ as
  \begin{equation}\label{EqnMLPStateFunctional}
    \mathbf{s}(\mathbf{x}) = [\bm{s}_1(\mathbf{x})^{\top}, \dots, \bm{s}_L(\mathbf{x})^{\top}]^{\top} .
  \end{equation}
\end{definition}
\noindent Let the total number of hidden neurons in $\mathbf{T}$ be $N = \sum_{l=1}^L n_l$. Denote $\mathbb{J} = \{1, 0\}$, and we have the state functional $\mathbf{s} \in \mathbb{J}^N$. Considering that a region in $\mathbb{R}^{n_0}$ corresponds to a realization of $\mathbf{s} \in \mathbb{J}^N$, it is clear that the maximal size of $\mathcal{R}$ is upper bounded by $2^N$. This gives us the following labeling scheme.
\begin{itemize}
  \item Any region $r \in \mathcal{R}$ corresponds to a unique element in $\mathbb{J}^N$; since $\mathbf{s}(\mathbf{x})$ is fixed for all $\mathbf{x} \in \mathcal{X}$ that fall in a same region $r$,  we use $\mathbf{s}(r) \in \mathbb{J}^N$ to label this region.
\end{itemize}
The following theorem from \cite{Montufar2014} gives a lower bound on the maximal size of $\mathcal{R}$.
\begin{theorem}[\cite{Montufar2014}]
  \label{TheoremMLPLinearRegionNumBound}
  For a ReLU based MLP $\mathbf{T}$ of $L$ hidden layers, whose layer widths satisfy $n_l \geq n_0$ for any $l \in \{1, \dots, L\}$,  the maximal size of $\mathcal{R}(\mathbf{T})$ is lower bounded by $\left( \prod_{l=1}^{L-1} \lfloor n_l/n_0 \rfloor^{n_0} \right) \sum_{j=0}^{n_0}{{n_L}\choose{j}}$, where $\lfloor\cdot\rfloor$ ignores the remainder. Assuming $n_1 = \cdots = n_L = n$, the lower bound has an order of $\mathcal{O}\left( (n/n_0)^{(L-1)n_0} n^{n_0}  \right)$.
\end{theorem}
\noindent The above theorem shows that the number of linear regions into which an MLP can partition the input space grows \emph{exponentially} with the network depth and \emph{polynomially} with the network width.
We have the following lemma adapted from \cite{Jia2019} to characterize the region-wise linear mappings.
\begin{lemma}[Linear Mapping of Region/Cell, an adaptation of Lemma 3.2 in \cite{Jia2019}]
  \label{LemmaRegionMapping}
  Given a ReLU based MLP $\mathbf{T}$ of $L$ hidden layers, for any region/cell $r\in \mathcal{R}(\mathbf{T})$, its associated linear mapping $\mathbf{T}^r$ is defined as
  \begin{align}
    \mathbf{T}^r     & = \prod\limits_{l=1}^{L}\mathbf{W}^r_l \label{EqnMLPRegionMapping}                   \\
    \mathbf{W}^r_{l} & = \textrm{diag}(\mathbf{s}_l(r))\mathbf{W}_{l} \label{EqnMLPRegionMappingLayerComp},
  \end{align}
  where $\textrm{diag}(\cdot)$ diagonalizes the state vector $\mathbf{s}_l(r)$.
\end{lemma}
\noindent Intuitively, the state vector $\mathbf{s}_l$ in (\ref{EqnMLPRegionMappingLayerComp}) selects a submatrix from $\mathbf{W}_{l}$ by setting those inactive rows as zero.

\subsection{Motivation to Connect the Local Linearities of the Two Worlds}

We implement the implicit surface field function $F$ by stacking on top of $\mathbf{T}$ a regression function $f: \mathbb{R}^{n_L} \rightarrow \mathbb{R}$, giving rise to a functional
\begin{displaymath}
  F(\mathbf{x}) = f\circ\mathbf{T}(\mathbf{x}) = \mathbf{w}_f^{\top}\mathbf{g}(\mathbf{W}_{L}\ldots \mathbf{g}(\mathbf{W}_1\mathbf{x})) ,
\end{displaymath}
where $\mathbf{w}_f \in \mathbb{R}^{n_L}$ is weight vector of the regressor. Since $F$ is defined in $\mathbb{R}^3$, we have $n_0 = 3$. Given that the MLP $\mathbf{T}$ partitions the input space $\mathbb{R}^3$ into a set $\mathcal{R}$ of linear regions, any region $r \in \mathcal{R}$ satisfies $\mathbf{x} \in \text{supp}(\mathbf{T}) \ \forall \ \mathbf{x} \in  r$, and can be uniquely indexed by its state vector $\mathbf{s}(r)$ defined by (\ref{EqnMLPStateFunctional}). For such a region $r$, we have the following corollary from Lemma \ref{LemmaRegionMapping} that characterizes the associated linear mappings defined at neurons of $\mathbf{T}$ and the final regressor.
\begin{corollary}\label{LemmaRegionAssociatedNeuronMapping}
  Given an implicit surface field function $F = f\circ\mathbf{T}$ built on a ReLU based MLP of $L$ hidden layers, for any $r\in \mathcal{R}(\mathbf{T})$, the associated neuron-wise linear mappings and that of the final regressor are defined as
  \begin{align}
    \mathbf{a}^r_{lk} & =
      \begin{cases}
        \pi_{k} \mathbf{W}_{l} \prod\limits_{i=1}^{l-1}\mathbf{W}^r_{i} & \text{when} \ l > 1 \\
        \pi_{k} \mathbf{W}_{l}                                          & \text{when} \ l = 1
      \end{cases} \label{EqnRegionAssociatedNeuronMapping}                                      \\
    \mathbf{a}_F^r    & = \mathbf{w}_f^{\top}\mathbf{T}^r = \mathbf{w}_f^{\top}\prod\limits_{i=1}^{L}\mathbf{W}^r_{i} \label{EqnSDFPlaneFunctional} ,
  \end{align}
  where $l \in \{1, \dots, L\}$, $k \in \{1, \dots, n_l\}$, and $\mathbf{T}^r$ and $\mathbf{W}^r_{i}$ are defined as in Lemma \ref{LemmaRegionMapping} (equations (\ref{EqnMLPRegionMapping}) and (\ref{EqnMLPRegionMappingLayerComp}) for $\mathbf{T}^r$ and $\mathbf{W}^r_l$, respectively).
\end{corollary}

Assume that the implicit function $F$ models a continuous and closed 2-dimensional surface manifold embedded in $\mathbb{R}^3$. Our problem of interest is to recover an explicit mesh $\mathcal{Z} = \{\mathcal{V}, \{\mathcal{P}\} \}$ from $F$, by extracting its zero-level isosurface $\{ \bm{x} \in \mathbb{R}^3 | F(\bm{x}) = 0 \}$. Section \ref{SecPolygonMesh} shows that $\mathcal{Z}$ has the piecewise linear parameterization of $\{ \bm{n}_{\mathcal{P}} \in \mathbb{R}^3 \}$. On the other hand, Corollary \ref{LemmaRegionAssociatedNeuronMapping} suggests that $F = f\circ \bm{T}$ in fact implements a piecewise linear function defined by the collection $\{ \mathbf{a}_F^r \in \mathbb{R}^3 | r \in \mathcal{R} \}$; consequently, the zero-level isosurface $\{ \bm{x} \in \mathbb{R}^3 | F(\bm{x}) = 0 \}$ can either be locally linear with polygon faces obtained by intersection with some of the linear regions $\{ r \in \mathcal{R} \}$, as illustrated in Fig. \ref{FigIntuition}, or in some special case coincide with hyperplane boundaries of some linear regions (detailed explanations are given in Section \ref{SecTheory}). We are interested in the former case and expect that the parameterization $\{ \bm{n}_{\mathcal{P}} \}$ of $\mathcal{Z}$ can be analytically identified as some of the liner mappings $\{ \mathbf{a}_F^r | r \in \mathcal{R} \}$. We prove in Section \ref{SecTheory} that this is indeed the case, and present an efficient algorithm of exactly meshing $\mathcal{Z}$ from $F$ in Section \ref{SecAlgorithm}, where we also show that $\bm{T}$ can be extended to incorporate shortcut connections and max pooling, which supports advanced architectures of $\bm{T}$. Details are presented as follows.


\section{Analytic Meshing from Deep Implicit Surface Networks}
\label{SecTheory}

\subsection{Analytic Cells and Analytic Faces Associated with a Deep Implicit Surface Network}
\label{SecAnalyticCellsFaces}

Corollary \ref{LemmaRegionAssociatedNeuronMapping} is useful to specify linear regions in $\mathcal{R}$ and the zero-level isosurface of the implicit function $F$. For any $r \in \mathcal{R}$, its boundary planes must be among the set
\begin{equation}\label{EqnBoundaryPlanes}
  \{ \bm{H}_{lk}^{r}\} \ \textrm{s.t.} \ \bm{H}_{lk}^{r} = \{ \mathbf{x} \in \mathbb{R}^3 | \mathbf{a}^{r}_{lk}\mathbf{x} = 0 \} ,
\end{equation}
where $l = 1, \dots, L$ and $k = 1, \dots, n_l$. The zero-level isosurface of $F$ in fact induces a set of region-associated planes in $\mathbb{R}^3$; the induced plane $\{\mathbf{x} \in \mathbb{R}^3 | \mathbf{a}_F^r\mathbf{x} = 0\}$ and the associated region $r$ have the following relations, assuming that the plane does not happen to coincide with any boundary plane of $\{ \bm{H}_{lk}^{r}\}$. For simplicity, we use the plane kernel/normal vector $\mathbf{a}_F^r$ to represent the region-associated plane induced by the zero-level isosurface of $F$.
\begin{itemize}
  \item \emph{Intersection} $\mathbf{a}_F^r$ splits the region $r$ into two halves, denoted as $r^{+}$ and $r^{-}$, such that $\forall \mathbf{x} \in r^{+}$, we have $\mathbf{a}_F^r\mathbf{x} > 0$ and $\forall \mathbf{x} \in r^{-}$, we have $\mathbf{a}_F^r\mathbf{x} \leq 0$.
  \item \emph{Non-intersection} We either have $\mathbf{a}_F^r\mathbf{x} > 0$ or $\mathbf{a}_F^r\mathbf{x} < 0$ for all $\mathbf{x} \in r$.
\end{itemize}
Let $\{\tilde{r} \in \widetilde{\mathcal{R}}\}$ denote the subset of regions in $\mathcal{R}$ that have the above relation of intersection. It is clear that the zero-level isosurface $\{ \mathbf{x} \in \mathbb{R}^3 | F(\mathbf{x}) = 0 \}$ defined on the support (\ref{EqnMLPSupport}) of $\mathbf{T}$ can be only in $\widetilde{\mathcal{R}}$.
Consider such a region $\tilde{r} \in \widetilde{\mathcal{R}}$; for any $\mathbf{x} \in \tilde{r}$, it must satisfy $(2 s_{lk}(\tilde{r}) - 1) \mathbf{a}^{\tilde{r}}_{lk}\mathbf{x} \geq 0$, which gives the following system of inequalities
\begin{equation}\label{EqnAnalyticCellSystem}
  (\mathbf{I} - 2\textrm{diag}(\mathbf{s}(\tilde{r})) \mathbf{A}^{\tilde{r}}\mathbf{x}  =
  \begin{bmatrix}
    (1 - 2s_{11}(\tilde{r}))\mathbf{a}^{\tilde{r}}_{11} \\
    \vdots                                              \\
    (1 - 2s_{lk}(\tilde{r}))\mathbf{a}^{\tilde{r}}_{lk} \\
    \vdots                                              \\
    (1 - 2s_{Ln_L}(\tilde{r}))\mathbf{a}^{\tilde{r}}_{Ln_L}
  \end{bmatrix}
  \mathbf{x} \preceq 0 ,
\end{equation}
where $\mathbf{I}$ is an identity matrix of compatible size, $\mathbf{A}^{\tilde{r}} \in \mathbb{R}^{N\times 3}$ collects the coefficients of the $N = \sum_{l=1}^L n_l$ inequalities, and the state functionals $s_{lk}$ and $\mathbf{s}$ are defined by (\ref{EqnNeuronStateFunctional}) and (\ref{EqnMLPStateFunctional}).
When the region is bounded, the system (\ref{EqnAnalyticCellSystem}) of inequalities essentially forms a convex polyhedral cell defined as
\begin{equation}\label{EqnAnalyticCell}
  \mathcal{C}_F^{\tilde{r}} = \{ \mathbf{x} \in \mathbb{R}^3 | (\mathbf{I} - 2\textrm{diag}(\mathbf{s}(\tilde{r})) \mathbf{A}^{\tilde{r}}\mathbf{x} \preceq 0 \} ,
\end{equation}
which we term as \emph{analytic cell of an implicit function's zero-level isosurface}, shortened as \emph{analytic cell}. We note that there could exist redundance in the defining inequalities of (\ref{EqnAnalyticCellSystem}); an analytic cell could also be a region open towards infinity in some directions.

Given the plane functional (\ref{EqnSDFPlaneFunctional}), we define the polygon face that is an intersection of analytic cell $\tilde{r}$ and zero-level isosurface of $F$ as
\begin{equation}\label{EqnAnalyticFace}
  \mathcal{P}_F^{\tilde{r}} = \{ \mathbf{x} \in \mathbb{R}^3 | \mathbf{a}_F^{\tilde{r}}\mathbf{x} = 0 , (\mathbf{I} - 2\textrm{diag}(\mathbf{s}(\tilde{r})) \mathbf{A}^{\tilde{r}}\mathbf{x} \preceq 0 \} ,
\end{equation}
which we term less precisely as \emph{analytic face of an implicit function's zero-level isosurface}, shortened as \emph{analytic face}, since it is possible that the face goes towards infinity in some directions. With the analytic form (\ref{EqnAnalyticFace}), we realize that a ReLU based MLP $F$ defines a piecewise planar zero-level isosurface, which could be an approximation to an underlying surface $\mathcal{M}$ when $F$ is trained using techniques presented in Section \ref{SecApps}.

\begin{figure}[!htbp]
  \vskip 0.1in
  \begin{center}
    \includegraphics[scale=0.55]{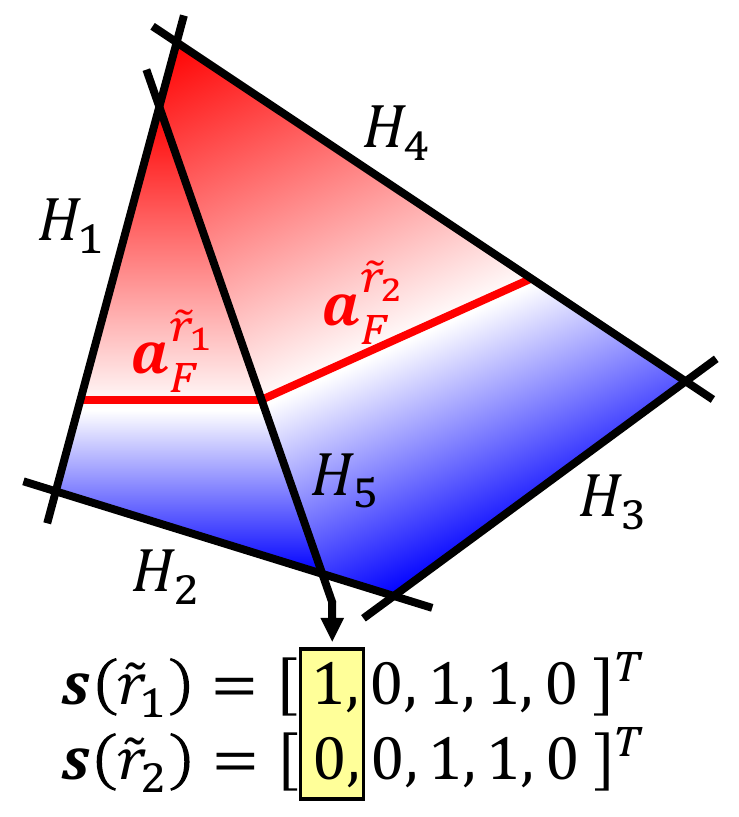} \vspace{-0.4cm}
    \caption{An 2D illustration for analytic cells and neural states.
      Red lines are the analytic faces.
      The states of two neighboring cells that share a boundary plane are switched at one neuron.
    }
    \label{FigAnalyticCellFaceIllus}
  \end{center}
  \vskip -0.1in
\end{figure}

\subsection{A Closed Mesh via Connected Analytic Faces in Analytic Cells}
\label{SecClosedMeshCondition}

We have stated in Section \ref{SecProbAndMotivation} that our problem of interest is to recover a surface mesh $\mathcal{Z}$ from the implicit function $F = f\circ \bm{T}$, and the surface of interest has the property of being continuous and closed. A closed, piecewise planar mesh $\mathcal{Z}$ means that every of its planar faces is connected with other faces via shared edges. Analysis in the preceding section shows that the zero-level isosurface of $F$ is piecewise planar whose associated analytic cells and analytic faces respectively satisfy (\ref{EqnAnalyticCell}) and (\ref{EqnAnalyticFace}). We have the following theorem that characterizes the conditions under which analytic faces (\ref{EqnAnalyticFace}) in their respective analytic cells (\ref{EqnAnalyticCell}) guarantee to connect and form a closed, piecewise planar $\mathcal{Z}$.
\begin{theorem}\label{MainResult}
  Assume that the zero-level isosurface $\mathcal{Z}$ of an implicit field function $F = f\circ\mathbf{T}$ defines a closed surface. If for any region/cell $r\in \mathcal{R}(\mathbf{T})$, its associated linear mapping $\mathbf{T}^r$ (\ref{EqnMLPRegionMapping}) and the induced plane $\mathbf{a}_F^r = \mathbf{w}_f^{\top}\mathbf{T}^r$ (\ref{EqnSDFPlaneFunctional}) are uniquely defined, i.e., $\mathbf{T}^r \neq \beta\mathbf{T}^{r'}$ and $\mathbf{a}_F^r \neq \beta\mathbf{a}_F^{r'}$ for any region pair of $r$ and $r'$, where $\beta$ is an arbitrary scaling, then analytic faces $\{ \mathcal{P}_F^{\tilde{r}} \}$ defined by (\ref{EqnAnalyticFace}) connect and exactly form the surface $\mathcal{Z}$ of polygon mesh.
\end{theorem}
\vspace{-0.2cm}

\begin{proof}
The proof is given in Appendix \ref{AppendixMainResultProof}. Given the assumed conditions, the proof can be sketched by first showing that each planar face on $\mathcal{Z}$ captured by the SDF $F = f\circ\mathbf{T}$ uniquely corresponds to an analytic face of an analytic cell, and then showing that for any pair of planar faces connected on $\mathcal{Z}$, their corresponding analytic faces are connected via boundaries of their respective analytic cells.
\end{proof}

We note that the conditions assumed in Theorem \ref{MainResult} can be practically met up to a numerical precision of the weights in $F = f\circ\mathbf{T}$. The proof also suggests an algorithm to identify the analytic polygon faces of the surface captured by $F$, which is to be presented in Section \ref{SecAlgorithm}.


\section{Practical and Efficient Implementations of the Analytic Marching Algorithm}
\label{SecAlgorithm}

In this section, we first present our proposed algorithm of \emph{analytic marching} for extraction of the piecewise planar, zero-level isosurface captured by an implicit $F = f\circ \bm{T}$; we then present its efficient implementations, including the CUDA version on GPUs, and customized strategies for triggering our algorithm when $F$ respectively represents an SDF or occupancy field. We finally discuss how the mesh obtained by analytic marching can be simplified, with least sacrifice of precision, to support efficient downstream processing.

\subsection{The Algorithm of Analytic Marching}
\label{SecBasicAM}

Given an implicit function $F = f\circ\mathbf{T}$ whose zero-level isofurface $\mathcal{Z} = \{ \mathbf{x} \in \mathbb{R}^3 | F(\mathbf{x}) = 0 \}$ defines a closed, piecewise planar surface, Theorem \ref{MainResult} suggests that obtaining the mesh $\mathcal{Z}$ concerns with identification of analytic faces $\{ \mathcal{P}_F^{\tilde{r}} | \tilde{r} \in \widetilde{\mathcal{R}} \}$ in analytic cells $\{\mathcal{C}_F^{\tilde{r}} | \tilde{r} \in \widetilde{\mathcal{R}} \}$. To this end, we propose an algorithm of \emph{analytic marching} that marches among $\{\mathcal{C}_F^{\tilde{r}} | \tilde{r} \in \widetilde{\mathcal{R}} \}$ to identify vertices and edges of the polygon faces, where the name is indeed to show respect to the classical discrete algorithm of marching cubes \cite{Lorensen1987}.

Specifically, analytic marching is triggered by identifying at least one point $\mathbf{x} \in \mathcal{Z}$ that satisfies $F(\mathbf{x}) = 0$; its state vector $\mathbf{s}(\mathbf{x})$ can be computed via (\ref{EqnMLPStateFunctional}), which specifies the analytic cell $\mathcal{C}_F^{\tilde{r}_{\mathbf{x}}}$ (\ref{EqnAnalyticCell}) and analytic face $\mathcal{P}_F^{\tilde{r}_{\mathbf{x}}}$ (\ref{EqnAnalyticFace}) where $\mathbf{x}$ resides. Analytic marching then successively solves a system of equations to analytically obtain vertices of the polygon face inside each analytic cell, and marches to neighboring analytic cells via transition of cell states. Details are given in Algorithm \ref{AlgAM}, with an illustration shown in Fig. \ref{FigAnalyticCellFaceIllus}.

\begin{algorithm}
    \caption{The algorithm of Analytic Marching}
    \label{AlgAM}
    \begin{flushleft}
    \textbf{INPUT:} An implicit surface field function $F = f \circ \bm{T}$ constructed from a ReLU based MLP \\
    \textbf{OUTPUT:} The exact zero-level isosurface of $F$ as a polygon mesh $\mathcal{Z} = \{\mathcal{V}, \{\mathcal{P}\} \}$
    \end{flushleft}

    \begin{algorithmic}[1]

    \State \algmultiline{%
        Initialize an active set $\mathcal{S}^{\bullet}=\emptyset$, and an inactive set $\mathcal{S}^{\circ}=\emptyset$.
    }

    \State \algmultiline{%
        Identify one point $\boldsymbol{x} \in \mathbb{R}^3$ that satisfies $ F(\boldsymbol{x}) = 0$ (and $\nabla_{\bm{x}} F(\bm{x}) \neq \boldsymbol{0} $).
    }

    \State \algmultiline{%
        Compute the state $\boldsymbol{s}(\boldsymbol{x})$ via (\ref{EqnMLPStateFunctional}), and push $\boldsymbol{s}(\boldsymbol{x})$ into $\mathcal{S}^{\bullet}$.
    }

    \While{$\mathcal{S}^{\bullet} \neq \emptyset$}
    \State Take an active state $\boldsymbol{s}(\tilde{r})$ from $\mathcal{S}^{\bullet}$ .

    \State \algmultiline{%
    Let $\mathcal{V}_{\mathcal{P}}^{\tilde{r}}$ be the set of defining vertices for the polygon face $\mathcal{P}_F^{\tilde{r}}$; enumerate all the pair $(\bm{H}_{lk}^{\tilde{r}}, \bm{H}_{l'k'}^{\tilde{r}})$ of boundary planes $\{ \bm{H}_{lk}^{\tilde{r}}\}$ defined by (\ref{EqnBoundaryPlanes}), with $l = 1, \dots, L$ and $k = 1, \dots, n_l$.
    }

    \State \algmultiline{%
    For each pair $(\bm{H}_{lk}^{\tilde{r}}, \bm{H}_{l'k'}^{\tilde{r}})$, together with $\mathcal{P}_F^{\tilde{r}}$, solve the following $3\times 3$ system of equations to have a $\bm{v} \in \mathbb{R}^3$
    }
    \begin{equation}\label{EqnVertexEquationSystem}
        [\mathbf{a}_{lk}^{\tilde{r}}; \mathbf{a}_{l'k'}^{\tilde{r}}; \mathbf{a}_F^{\tilde{r}}] \mathbf{x} = \bm{0}  .
    \end{equation}

    \State \algmultiline{%
    Confirm the validity of $\mathbf{v} \in \mathcal{V}_{\mathcal{P}}^{\tilde{r}}$ when it satisfies the boundary condition (\ref{EqnAnalyticCellSystem}) of the cell $\mathcal{C}_F^{\tilde{r}}$.
    }

    \State \algmultiline{%
    Form $\mathcal{V}_{\mathcal{P}}^{\tilde{r}}$ of the face $\mathcal{P}_F^{\tilde{r}}$ with all the valid vertices obtained by solving (\ref{EqnVertexEquationSystem}); push vertices of $\mathcal{V}_{\mathcal{P}}^{\tilde{r}}$ into $\mathcal{V}$, and the face $\mathcal{P}_F^{\tilde{r}}$ into $\mathcal{Z}$.
    }

    \State \algmultiline{%
    Record all the boundary planes $\{ \widehat{\bm{H}}_{lk}^{\tilde{r}} \}$ of $\mathcal{C}_F^{\tilde{r}}$ that give valid vertices; for each $\widehat{\bm{H}}_{lk}^{\tilde{r}}$,  infer the state $\bm{s}(\tilde{r}_{\textrm{\tiny Neighbor}})$ of a neighboring analytic cell by switching $\boldsymbol{s}(\tilde{r})$ at $s(\tilde{r}_{lk})$.
    }

    \State \algmultiline{%
        Push $\boldsymbol{s}(\tilde{r})$ out of the active set $\mathcal{S}^{\bullet}$ and into the inactive set $\mathcal{S}^{\circ}$; push $\{ \bm{s}(\tilde{r}_{\textrm{\tiny Neighbor}}) | \bm{s}(\tilde{r}_{\textrm{\tiny Neighbor}}) \not\in  \mathcal{S}^{\circ} \}$ activated in the preceding step into the active set $\mathcal{S}^{\bullet}$.
    }

    \EndWhile

    \end{algorithmic}
\end{algorithm}

Given $F$ and an arbitrary space point, the zero-crossing point $\bm{x}$ in Step 2 can be obtained simply by solving the following problem via stochastic gradient descent (SGD)
\begin{equation}\label{EqnSurfacePointOptim}
  \min_{\mathbf{x} \in \mathbb{R}^3} | F(\mathbf{x}) | .
\end{equation}
In practice, it is not necessary for the obtained $\bm{x}$ to exactly satisfy $F(\boldsymbol{x}) = 0$; the algorithm works as long as $F(\boldsymbol{x})$ is sufficiently small that ensures $\bm{x}$ falls in an analytic cell.

\vspace{0.1cm}
\noindent\textbf{Algorithmic guarantee} Theorem \ref{MainResult} guarantees that when the zero-level isosurface $\mathcal{Z}$ of the implicit function $F = f \circ\bm{T}$ is closed, identification of all the analytic faces forms the closed surface mesh. The proposed analytic marching algorithm is anchored at cell state transition whose success is of high probability due to a phenomenon similar to the blessing of dimensionality \cite{Gorban2018} --- it is of low probability that edges connecting planar faces of $\mathcal{Z}$ coincide with edges of analytic cells (cf. proof of Theorem \ref{MainResult} for detailed analysis).

\subsubsection{Analysis of Computational Complexity}
\label{SecComplexityAnalysis}

Consider the implicit function $F = f\circ \mathbf{T}$ built on an MLP of $L$ hidden layers, each of which has $n_l$ neurons, $l = 1, \dots, L$. Let $N = n_1 + \dots, n_L$. For ease of analysis, we assume $n_1 = \cdots = n_L = n$ and thus $N = nL$. The computations inside each analytic cell concern with computing the boundary planes, solving a maximal number of ${N}\choose{2}$ equation systems (\ref{EqnVertexEquationSystem}), and checking the validity of resulting vertices, which give a complexity order of $\mathcal{O}(n^3L^3)$. We know from \cite{Montufar2014} that the maximal size of the set $\mathcal{R}(\mathbf{T})$ of linear regions in general has an order of $\mathcal{O}\left( (n/n_0)^{(L-1)n_0} n^{n_0}  \right)$, where $n_0$ is the dimensionality of input space. Since our focus of interest is the 2-dimensional surface embedded in the 3D space, we have $n_0 = 2$ and thus the maximal size of $\mathcal{R}(\mathbf{T})$, which bounds the maximal number of analytic cells, in general has an order of $\mathcal{O}\left( (n/2)^{2(L-1)} n^2  \right)$. Overall, the complexity of our analytic marching algorithm has an order of $\mathcal{O}\left( (n/2)^{2(L-1)} n^5L^3 \right)$, which is exponential w.r.t. the MLP depth $L$ and polynomial w.r.t. the MLP width $n$. Note that the result can be further improved by applying the efficient implementations to be presented in Section \ref{SecEfficientImplementations}.

The above analysis shows that the complexity nature of analytic marching is the complexity of implicit function, which is completely different from those of existing algorithms, such as marching cubes \cite{Lorensen1987}, whose complexities are irrelevant to function complexities but rather depend on the discretized resolutions of the 3D space. Our algorithm thus provides an opportunity to recover highly precise mesh reconstruction by using networks of low complexities.

\subsection{Efficient Implementations}
\label{SecEfficientImplementations}

\subsubsection{Pivoting Enumeration in a Working Cell}

Steps 6 - 11 in Algorithm \ref{AlgAM} involve enumeration of all the pairs from the boundary planes $\{ \bm{H}_{lk}^{\tilde{r}}\}$, with $l = 1, \dots, L$ and $k = 1, \dots, n_l$, in order to find the valid vertices $\{ \bm{v} \in \mathcal{V}_{\mathcal{P}}^{\tilde{r}} \}$ for the polygon face $\mathcal{P}_F^{\tilde{r}}$ in the working cell $\mathcal{C}_F^{\tilde{r}}$. The process is less efficient and requires a postprocessing step to identify the traversal order of valid vertices. To improve the efficiency, we present a pivoting-based enumeration scheme \cite{David1991} whose details are as follows.

Pivoting enumeration starts with identifying a point $\boldsymbol{x} \in \mathcal{P}_F^{\tilde{r}}$ and a hyperplane $\bm{H}_{l k}^{\tilde{r}}$, among $\{ \bm{H}_{lk}^{\tilde{r}}\}$, that is a true boundary of the working cell $\mathcal{C}_F^{\tilde{r}}$; we present how such an $\boldsymbol{x}$ and $\bm{H}_{l k}^{\tilde{r}}$ can be identified shortly. The scheme then establishes an index set $\mathcal{I}^{\tilde{r}} = \{ (l_1, k_1), \dots, (l_{N}, k_{N}) \}$ by sorting, in an increasing order, the Euclidean distances from each of the $N$ hyperplanes in $\{ \bm{H}_{lk}^{\tilde{r}}\}$ to the point $\boldsymbol{x} \in \mathcal{P}_F^{\tilde{r}}$, where $N = n_1 + \dots, + n_L$ is the total number of neurons in $\bm{T}$. Introduce the auxiliary $t, t', t'' \in \mathbb{N}$, and we use $(l_t, k_t)$ to index the boundary plane $\bm{H}_{l k}^{\tilde{r}}$ identified in the beginning, written as $\bm{H}_{l_t k_t}^{\tilde{r}}$; for a later reference, we also use $(l_{*}, k_{*})$ and $\bm{H}_{l_{*} k_{*}}^{\tilde{r}}$ to refer to this same boundary. Let $\mathcal{I}_{/_t}^{\tilde{r}} = \{ (l_1, k_1), \dots, (l_{t-1}, k_{t-1}), (l_{t+1}, k_{t+1}), \dots, (l_{N}, k_{N}) \}$,
and we have $| \mathcal{I}_{/_t}^{\tilde{r}} | = N - 1$.
By initializing $\mathcal{V}_{\mathcal{P}}^{\tilde{r}} = \emptyset$ and $t'' = 1$, the scheme firstly repeats the following steps: 1) solves the system of equations $[\mathbf{a}_{l_{t} k_{t}}^{\tilde{r}}; \mathbf{a}_{l_{t''} k_{t''}}^{\tilde{r}}; \mathbf{a}_F^{\tilde{r}}] \mathbf{x} = \bm{0}$ to have a vertex candidate $\bm{v} \in \mathbb{R}^3$; 2) confirms the validity of $\bm{v} \in \mathcal{V}_{\mathcal{P}}^{\tilde{r}}$ by checking whether it satisfies the boundary condition (\ref{EqnAnalyticCellSystem}) of the cell $\mathcal{C}_F^{\tilde{r}}$; 3) if the above step is true, pushes $\bm{v}$ into $\mathcal{V}_{\mathcal{P}}^{\tilde{r}}$, lets $(l_{t'}, k_{t'}) = (l_{t}, k_{t})$ and updates $(l_t, k_t) = (l_{t''}, k_{t''})$, and then \emph{exits}; otherwise updates $t'' \leftarrow t'' + 1$ upon $(l_{t''+1}, k_{t''+1}) \in \mathcal{I}_{/_t}^{\tilde{r}}$ or $t'' \leftarrow t'' + 2$ upon $(l_{t''+2}, k_{t''+2}) \in \mathcal{I}_{/_t}^{\tilde{r}}$, and goes back to step 1. Given that $\bm{H}_{l_t k_t}^{\tilde{r}}$ is a true boundary of the polyhedral cell $\mathcal{C}_F^{\tilde{r}}$, it is guaranteed for the above steps to find a valid  $\bm{v} \in \mathcal{V}_{\mathcal{P}}^{\tilde{r}}$. Let $\mathcal{I}_{/_{t'}/_{t}}^{\tilde{r}} = \{ (l_1, k_1), \dots, (l_{t'-1}, k_{t'-1}),$ $(l_{t'+1}, k_{t'+1}), \dots, (l_{t-1}, k_{t-1}), (l_{t+1}, k_{t+1}), \dots, (l_{N}, k_{N}) \}$ and we have $| \mathcal{I}_{/_{t'}/_{t}}^{\tilde{r}} | = N-2$, where we assume $t' < t$ for notational simplicity, the scheme then repeats the following steps: 1) solves the system of equations $[\mathbf{a}_{l_{t} k_{t}}^{\tilde{r}}; \mathbf{a}_{l_{t''} k_{t''}}^{\tilde{r}}; \mathbf{a}_F^{\tilde{r}}] \mathbf{x} = \bm{0}$ to have a vertex candidate $\bm{v} \in \mathbb{R}^3$; 2) confirms the validity of $\bm{v} \in \mathcal{V}_{\mathcal{P}}^{\tilde{r}}$ by checking whether it satisfies the boundary condition (\ref{EqnAnalyticCellSystem}) of the cell $\mathcal{C}_F^{\tilde{r}}$; 3) if the above step is true, pushes $\bm{v}$ into $\mathcal{V}_{\mathcal{P}}^{\tilde{r}}$, and either \emph{exits} when $(l_{t''}, k_{t''})$ indexes the same boundary plane as the starting one $\bm{H}_{l_{*} k_{*}}^{\tilde{r}}$, or \emph{sequentially} updates $(l_{t'}, k_{t'}) = (l_{t}, k_{t})$, $(l_t, k_t) = (l_{t''}, k_{t''})$, and $\mathcal{I}_{/_{t'}/_t}^{\tilde{r}} = \{ (l_1, k_1), \dots, (l_{t'-1}, k_{t'-1}), (l_{t'+1}, k_{t'+1}), \dots, (l_{t-1}, k_{t-1}), $ $(l_{t+1}, k_{t+1}), \dots, (l_{N}, k_{N}) \}$, and (re-)sets $t'' = 1$; otherwise updates $t'' \leftarrow t'' + 1$ upon $(l_{t''+1}, k_{t''+1}) \in \mathcal{I}_{/_{t'}/_t}^{\tilde{r}}$ or $t'' \leftarrow t'' + 2$ upon $(l_{t''+2}, k_{t''+2}) \in \mathcal{I}_{/_{t'}/_t}^{\tilde{r}}$, and goes back to step 1. The ending condition in the above step 3 suggests that the scheme has circled back and found all the valid vertices $\{ \bm{v} \in \mathcal{V}_{\mathcal{P}}^{\tilde{r}} \}$ that form the polygon face $\mathcal{P}_F^{\tilde{r}}$. We summarize the algorithm in Appendix \ref{AppendixPivotingEnum}.

Sorting planes according to Euclidean distances reduces the number of iterations required to find the valid vertices. We note that acquiring the starting boundary plane $\bm{H}_{l_{*} k_{*}}^{\tilde{r}}$ and point $\boldsymbol{x} \in \mathcal{P}_F^{\tilde{r}}$ does not require any extra effort; $\bm{H}_{l_{*} k_{*}}^{\tilde{r}}$ can be set exactly as the switching plane that gives the current cell state (cf. Step 10 of Algorithm \ref{AlgAM}), and $\boldsymbol{x}$ can be set as the middle point of the switching edge on the switching plane. The presented pivoting enumeration has an average-case complexity of
$\mathcal{O}\left( |\mathcal{V}_{\mathcal{P}}| n^2L^2 \right)$, where $|\mathcal{V}_{\mathcal{P}}|$ is the average number of vertices per analytic face; it reduces the overall complexity of analytic marching to an order of $\mathcal{O}\left( (n/2)^{2(L-1)} |\mathcal{V_{\mathcal{P}}}|n^4L^2 \right)$.

\subsubsection{Parallel Marching with CUDA Implementation}

Our proposed analytic marching naturally supports parallel implementation. Instead of starting from a single initial point $\boldsymbol{x} \in \mathcal{Z}$ obtained by solving (\ref{EqnSurfacePointOptim}) (Step 2 of Algorithm \ref{AlgAM}), one may initialize as many of such points as possible in parallel; the algorithm can then be fully parallelized by simultaneously marching towards all the analytic cells in the active set $\mathcal{S}^{\bullet}$ that are unsolved to get their respectively analytic faces. \emph{Parallel marching} would improve the efficiency of analytic marching significantly.

Parallel marching can be practically achieved on parallel computing devices (e.g. GPUs). As a contribution to the community, we present a CUDA implementation of analytic marching algorithm to support parallel marching. The implementation is incorporated in the package of \texttt{AnalyticMesh} publicly available at \url{https://github.com/Karbo123/AnalyticMesh}. With CUDA implementation on Nvidia GPUs, the efficiency of analytic marching can be improved at an order of 10. We report empirical running time comparisons in Section \ref{SecExpsAnalysisAM}.

\subsection{Customized Schemes for Triggering the Algorithm}
\label{SecCustomizedInitialization}

The algorithm of analytic marching can be triggered by solving (\ref{EqnSurfacePointOptim}) via SGD to find $\bm{x} \in \mathcal{Z}$. Depending on whether the implicit function $F$ is an SDF or an occupancy field (OF) function, one may leverage the defining properties of the respective fields to have customized triggering strategies. In this section, we present two such schemes respectively specialized for SDF and OF. Empirical results in Section \ref{SecExpsAnalysisAM} confirm the efficiency of the two schemes when compared with triggering by solving (\ref{EqnSurfacePointOptim}) via SGD.

\subsubsection{Sphere Tracing-based Triggering for SDF}

By utilizing the property that SDF satisfies the Eikonal equation $| \nabla_{\bm{x}} F | = 1$, sphere tracing \cite{Hart1996} greatly accelerates the rendering of implicit surface. This inspires us to have an efficient sphere tracing-based scheme to find $\bm{x} \in \mathcal{Z}$ for SDF. More specifically, let $\boldsymbol{x}_0 \in \mathbb{R}^3$ denote an arbitrary initial point in the implicit field of SDF $F$; given an $\bm{x}_t \in \mathbb{R}^3$ at a time step $t$, we use the following updating rule to have $\bm{x}_{t+1}$
\begin{equation}
  \label{EqnSphereTracing}
  \boldsymbol{x}_{t+1} \leftarrow \boldsymbol{x}_{t} - \eta F(\boldsymbol{x}_t) \nabla_{\bm{x}_t} F ,
\end{equation}
where $\eta \leq 1$ is the step size. Given that $F(\boldsymbol{x}_t)$ computes the signed distance to the surface $\mathcal{Z}$ at $\bm{x}_t$, the rule (\ref{EqnSphereTracing}) thus accelerates standard SGD updating when $\bm{x}_t$ is far away from $\mathcal{Z}$, and it has a damping effect to prevent surface penetration when $\bm{x}_t$ moves very close to $\mathcal{Z}$.

\subsubsection{Dichotomy-based Triggering for Occupancy Field}

A binary OF is usually relaxed by using a sigmoid function $f$ to construct $F = f\circ \bm{T}$. As such, gradient-based optimization (e.g. SGD) is less effective to find $\bm{x} \in \mathcal{Z}$ from an arbitrary field point, especially when the initial point is far away from the surface. Fortunately, Bolzano's theorem \cite{Russ1980} states that a continuous function has a root in an interval if it has values of opposite signs inside that interval. Based on this, we propose a dichotomy-based scheme to find $\bm{x} \in \mathcal{Z}$ for OF. More specifically, we first randomly sample seed points in the implicit OF until a pair $\{\boldsymbol{x}_0^{+}, \boldsymbol{x}_0^{-}\}$ is obtained which satisfies $F(\bm{x}_0^{+}) > 0$ and $F(\bm{x}_0^{-}) < 0$. According to Bolzano's theorem, we can assert that there must be at least one zero-crossing point $\bm{x}$ satisfying $F(\boldsymbol{x}) = 0$ on the line segment $\bm{x}_0^{+} - \bm{x}_0^{-}$. To find a zero-crossing point, we repeatedly bisect the line segment, and then select the subinterval at an iteration $t$ whose two end points $\bm{x}_t^{+}$ and $\bm{x}_t^{-}$ have opposite signs of occupancy. The process continues until $F(\bm{x}_t^{+}) - F(\bm{x}_t^{-}) \leq \epsilon$ for a specified tolerance $\epsilon$.

\subsection{Postprocessing for Mesh Simplification}
\label{SecMeshSimplification}

The polygon mesh $\mathcal{Z}$ obtained by analytic marching is an exact solution of the zero-level isosurface of an implicit function $F = f \circ \bm{T}$. When the network $\bm{T}$ is large, there would be a huge number of polygon faces in the obtained $\mathcal{Z}$ (cf. Table \ref{TableExpsSimplificationNumerics} for a reference of the number of faces practically obtained). This would bring inconvenience for subsequent processing on mesh, e.g., rendering and texturing. To reduce the number of faces with least sacrifice of mesh precision, we adopt the quadric edge collapse decimation (QECD) algorithm \cite{Garland1997} to simplify the obtained $\mathcal{Z}$; QECD estimates the placement of vertices of collapsed edges by minimizing the distances to neighboring faces, and is hence able to preserve sharp features of the original $\mathcal{Z}$.
In case that the implicit function $F = f \circ \bm{T}$ itself captures a less ideal surface $\mathcal{Z}$, e.g., a rugged or non-watertight surface, our analytic marching would also produce the exact but less desirable $\mathcal{Z}$. One could apply postprocessing steps, such as smoothing \cite{Guennebaud2007} or holes filling \cite{Liepa2003}, to improve the visual quality.
We also note that our obtained polygon meshes can be easily converted as triangular ones, simply by subdividing polygons into triangles along diagonals.

\begin{figure}[!htbp]
  \vskip 0.1in
  \begin{center}
    \includegraphics[scale=0.25]{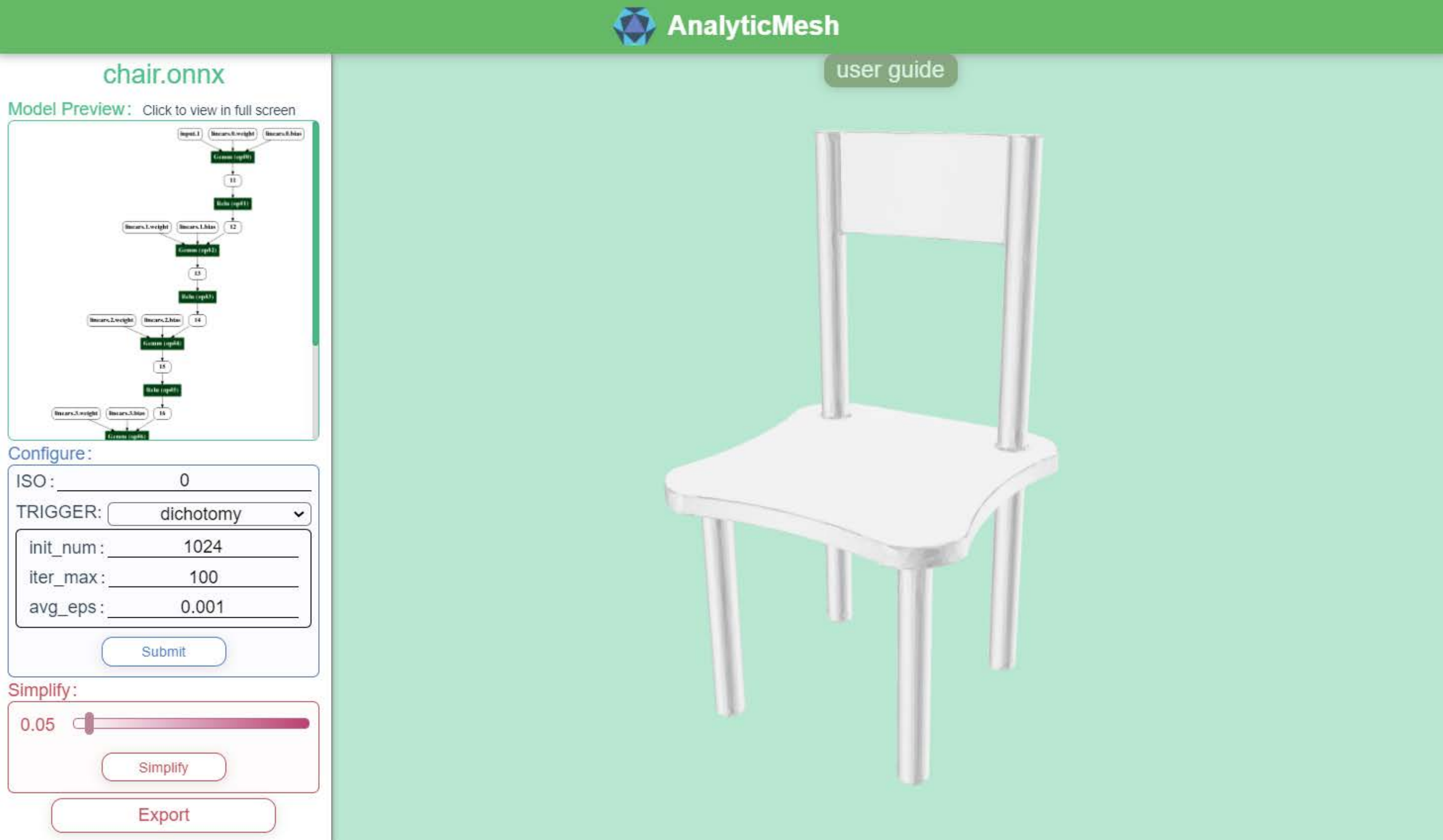} \vspace{-0.3cm}
    \caption{The user interface that facilitates use of the publicly released \texttt{AnalyticMesh} package.
    }
    \label{FigSoftwareInterface}
  \end{center}
  \vskip -0.1in
\end{figure}

We incorporate the above postprocessing operations into the publicly released package \texttt{AnalyticMesh}, in which we also provide a handle to control the number of polygon faces. Fig. \ref{FigSoftwareInterface} shows the interface. Experiments in Section \ref{SecExp} confirm that the proposed postprocessing is able to effectively simplify the meshes obtained by analytic marching; in many cases, it produces visually more pleasant results.

\section{Extensions to Other Architectures}
\label{SecExtOtherArchitecture}

We have so far focused our analysis on implicit functions constructed from MLPs with ReLU activations. In this section, we show that our proposed analytic meshing is applicable to implicit functions constructed from more advanced architectures, including those with shortcut connections \cite{He2016} and max pooling. These architectures are used in some of the recent deep learning surface reconstruction methods \cite{Park2019,Chen2018,Chen2019,Tretschk2020}. To facilitate the discussion, we will override some of the previously introduced math notations, which are self clear in the respective contexts.

\begin{figure}[!htbp]
  \vskip 0.1in
  \begin{center}
    \includegraphics[scale=0.4]{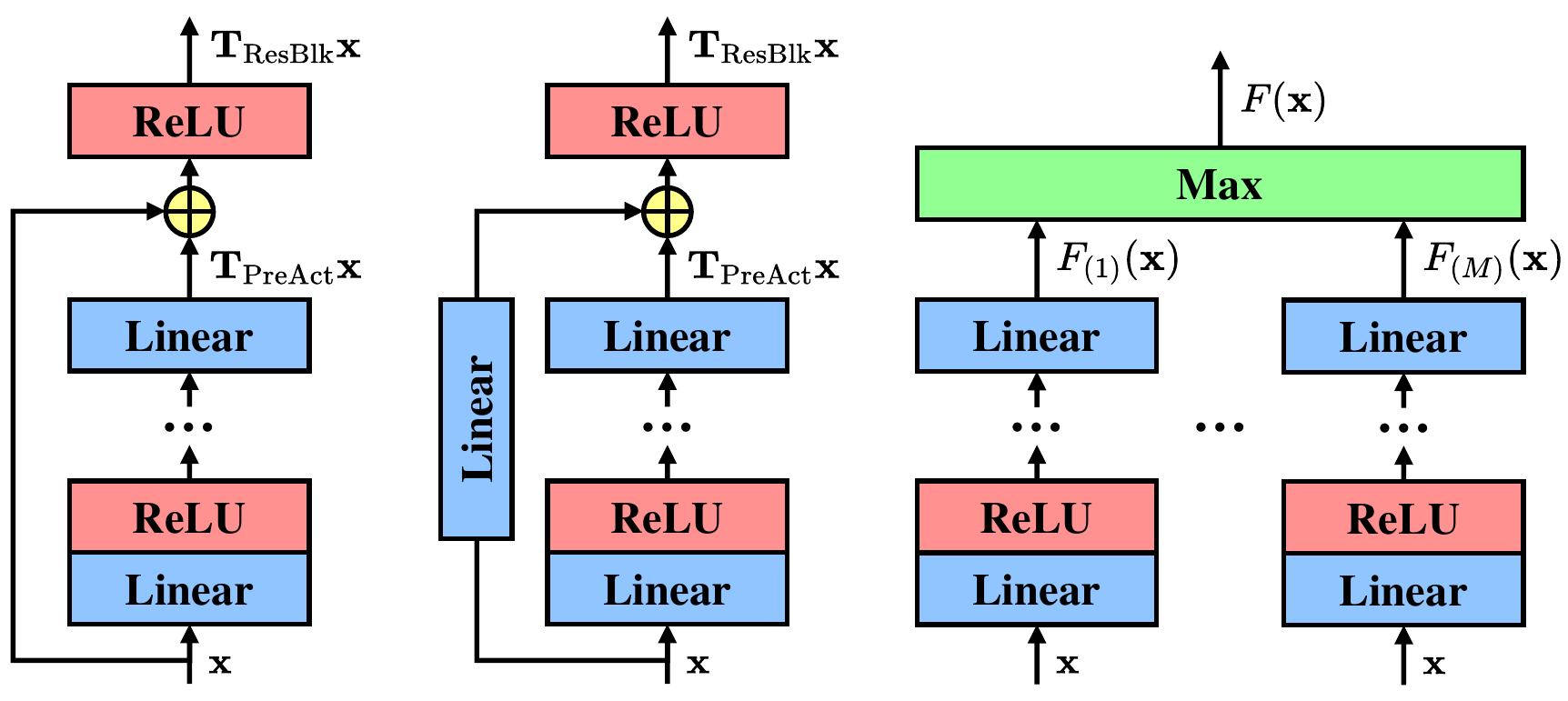} 

    \scalebox{1.0}
    {
      \begin{tabular}{C{1.8cm} C{1.8cm} C{3.6cm}}
      \ \ \ \ \ \ \ \ \  (a)  \ \ \ \ \  & \ \ \ \ \ \ (b) &  (c) \\
      \end{tabular}
    }
    \vspace{-0.3cm}
    \caption{Illustration of the three network architectures discussed in Sections \ref{SecMLPwithShortcuts} and \ref{SecFinalMaxPoolingTheory}.
      (a) A residual block with a shortcut connection.
      (b) A residual block with a shortcut connection of linear mapping.
      (c) A network with max pooling as the final aggregation of the outputs from multiple subnetworks.
    }
    \label{FigExtOtherArchitecture}
  \end{center}
  \vskip -0.1in
\end{figure}

\subsection{Multi-Layer Perceptrons with Shortcut Connections}
\label{SecMLPwithShortcuts}

We consider two prototypical residual blocks with shortcut connections to present the extension. Fig. \ref{FigExtOtherArchitecture} gives the illustration.
The first residual block aggregates two paths of forward signal propagation before a final ReLU activation, where one path is a shortcut connection and the other path stacks $L$ hidden layers respectively of $n_l$ neurons, $l \in \{1, \dots, L\}$. Denote an input $\bm{x} \in \mathcal{X} \subset \mathbb{R}^{n_0}$; the residual block thus defines a mapping $\bm{T}_{\textrm{\tiny ResBlk}}\bm{x} = \bm{g}(\bm{x} + \mathbf{T}_{\textrm{\tiny PreAct}}\bm{x})$, with the mapping $\mathbf{T}_{\textrm{\tiny PreAct}}\bm{x} = \mathbf{W}_L\bm{g}(\ldots \mathbf{g}(\mathbf{W}_1\bm{x}))$, which also implies $n_0 = n_L$. Analysis in Section \ref{SecLocalLinearityofMLP} shows that $\bm{T}_{\textrm{\tiny PreAct}}$ in fact partitions the space $\mathcal{X}$ into a number of linear regions. To understand how the residual block $\bm{T}_{\textrm{\tiny ResBlk}}$ partitions the space, we first note from Definition \ref{DefinitionNeuronMLPState} that the state functional of layer $l$ in $\bm{T}_{\textrm{\tiny PreAct}}$, $l \in \{1, \dots, L-1\}$, is $\mathbf{s}_l(\mathbf{x}) = [s_{l1}(\mathbf{x}), \dots, s_{ln_l}(\mathbf{x})]^{\top}$, where $s_{lk}(\mathbf{x})$, $k \in \{1, \dots, n_l\}$, is defined by (\ref{EqnNeuronStateFunctional}); we then extend Definition \ref{DefinitionNeuronMLPState} and define the state functional for the output of the final layer of $\bm{T}_{\textrm{\tiny ResBlk}}$ as
\begin{equation}\label{EqnResBlockStateFunctional}
  \mathbf{s}_L(\mathbf{x}) = [s_{L1}(\mathbf{x}), \dots, s_{Ln_L}(\mathbf{x})]^{\top} ,
\end{equation}
\begin{equation}
  \textrm{s.t.} \ s_{Lk}(\mathbf{x}) =
  \begin{cases}
    1 & \text{if $\pi_k (\bm{x} + \mathbf{T}_{\textrm{\tiny PreAct}}\bm{x}) > 0$}      \\
    0 & \text{if $\pi_k (\bm{x} + \mathbf{T}_{\textrm{\tiny PreAct}}\bm{x}) \leq 0$} ,
  \end{cases} \nonumber
\end{equation}
where $k \in \{1, \dots, n_L\}$. We thus have the state functional of $\bm{T}_{\textrm{\tiny ResBlk}}$ as $\bm{s}{\textrm{\tiny ResBlk}}(\bm{x}) = [\bm{s}_1(\mathbf{x})^{\top}, \dots, \bm{s}_L(\mathbf{x})^{\top}]^{\top}$. Let $\mathcal{R}_{\textrm{\tiny ResBlk}}$ denote the set of linear regions in $\mathbb{R}^{n_0}$ that are partitioned by $\bm{T}_{\textrm{\tiny ResBlk}}$. With definition (\ref{EqnResBlockStateFunctional}), we can label any region $r \in \mathcal{R}_{\textrm{\tiny ResBlk}}$ as $\bm{s}{\textrm{\tiny ResBlk}}(r) \in \mathbb{J}^N$, where $N = \sum_{l=1}^L n_l$. The region-wise linear mapping analogous to Lemma \ref{LemmaRegionMapping} can thus be defined as
\begin{align}
  \mathbf{T}_{\textrm{\tiny ResBlk}}^r & =  \textrm{diag}(\mathbf{s}_L(r)) \left(\bm{I} + \bm{W}_L \prod\limits_{l=1}^{L-1} \mathbf{W}^r_l \right) \label{EqnResBlkRegionMapping} \\
  \textrm{s.t.} \ \mathbf{W}^r_{l}     & = \textrm{diag}(\mathbf{s}_l(r))\mathbf{W}_{l} \ \ l \in \{1, \dots, L-1 \} , \nonumber
\end{align}
where $\bm{I}$ is an identity matrix of compatible size. Consequently, the neuron-wise linear mapping analogous to Corollary \ref{LemmaRegionAssociatedNeuronMapping} is defined as
\begin{equation}\label{EqnRegionAssociatedNeuronMapping4ResBlock}
  \mathbf{a}^r_{lk} =
  \begin{cases}
    \pi_{k} \left(\bm{I} + \bm{W}_L \prod\limits_{i=1}^{L-1} \mathbf{W}^r_i \right)  & \text{when $l = L$}                     \\
    \pi_{k} \left( \bm{W}_l \prod\limits_{i=1}^{l-1} \bm{W}_i^r \right)                     & \text{when $l \in \{2, \dots, L-1\}$}                                          \\
    \pi_{k} \bm{W}_l                                                                 & \text{when $l = 1$} ,
  \end{cases}
\end{equation}
where $k \in \{1, \dots, n_l\}$ and $\mathbf{W}^r_{i}$ is defined the same as $\mathbf{W}^r_{l}$ in (\ref{EqnResBlkRegionMapping}). One may use $\bm{T}_{\textrm{\tiny ResBlk}}$ as building blocks to construct an MLP $\bm{T}$ with shortcut connections, and consequently an implicit function $F = f\circ \bm{T}$. Given the definitions (\ref{EqnResBlockStateFunctional}), (\ref{EqnResBlkRegionMapping}), and (\ref{EqnRegionAssociatedNeuronMapping4ResBlock}), the theoretical analysis in Section \ref{SecTheory} and the analytic marching algorithm in Section \ref{SecAlgorithm} can be readily applied to such an $F$. This extends our proposed method to architectures incorporating $\bm{T}_{\textrm{\tiny ResBlk}}$ as building blocks.

The second residual block simply replaces the path of shortcut connection with a linear mapping $\bm{V}\bm{x}$, giving rise to a mapping of the residual block as $\bm{T}_{\textrm{\tiny ResBlk}}\bm{x} = \bm{g}(\bm{V}\bm{x} + \mathbf{T}_{\textrm{\tiny PreAct}}\bm{x})$, where $\bm{V} \in \mathbb{R}^{n_L\times n_0}$ and $\mathbf{T}_{\textrm{\tiny PreAct}}$ is the same as for the first residual block. This second residual block supports $n_0 \neq n_L$. Definitions similar to (\ref{EqnResBlockStateFunctional}), (\ref{EqnResBlkRegionMapping}), and (\ref{EqnRegionAssociatedNeuronMapping4ResBlock}) can be derived correspondingly; this extends our method to architectures incorporating the second type of residual blocks as building blocks.

\subsection{Max pooling as a Final Aggregation of Deep Implicit Surface Networks}
\label{SecFinalMaxPoolingTheory}

A few recent methods \cite{Chen2019,Tretschk2020} model a solid surface using constructive solid geometry \cite{Botsch2010}. They technically implement a union operation via a final max pooling over multiple deep implicit surface (sub-)networks. Fig. \ref{FigExtOtherArchitecture} gives an illustration. In this section, we show that our proposed method can also be extended to achieve analytic meshing from such a union of subnetworks.

Let $F_{(i)} = f_{(i)} \circ \bm{T}_{(i)}$, $i = 1, \dots, M$, denote the individual implicit functions constructed from $M$ subnetworks respectively of $L_{(i)}$ hidden layers. Each $F_{(i)}$ takes as input a same $\bm{x} \in \mathcal{X} \subset \mathbb{R}^{n_0}$; $n_0 = 3$ in the context of interest. Their union via max pooling computes
\begin{equation}\label{EqnMaxPoolingForwardPropagation}
  F(\bm{x}) = \max_{i \in \{1, \ldots, M\}}  F_{(i)} (\bm{x}) .
\end{equation}
Analysis in Sections \ref{SecProbAndMotivation} and \ref{SecTheory} suggests that each $\bm{T}_{(i)}$ partitions the input space $\mathcal{X}$ into a set $\mathcal{R}_{(i)}$ of linear regions/cells (convex polyhedrons), of which $\widetilde{\mathcal{R}}_{(i)}$ is the set of analytic cells relevant to the zero-level isosurface $\{ \bm{x} \in \mathbb{R}^3 | F_{(i)}(\bm{x}) = 0 \}$. For any $\bm{x} \in \mathcal{X}$, without loss of generality we assume that it falls in the cells $r_{(i)} \in \mathcal{R}_{(i)}$, $i = 1, \dots, M$, respectively partitioned by the $M$ subnetworks. Given the neuron-wise linear mappings defined in Corollary \ref{LemmaRegionAssociatedNeuronMapping}, we can spell out (\ref{EqnMaxPoolingForwardPropagation}) for a local region around $\bm{x}$ as
\begin{equation}\label{EqnMaxPoolingForwardPropagationLocal}
  F(\bm{x}) = \max\{ \bm{a}_{F_{(1)}}^{r_{(1)}}\bm{x}, \dots, \bm{a}_{F_{(M)}}^{r_{(M)}}\bm{x} \} \ \forall \ \bm{x} \in r_{(1)} \cap \cdots \cap r_{(M)} ,
\end{equation}
which is a point-wise maximum of linear functions. Let $r := r_{(1)} \cap \cdots \cap r_{(M)} \subset \mathbb{R}^3$;
classical linear algebra suggests that $r$ is a convex polyhedron as well.

To achieve analytic meshing from $r$, the key is to identify (possibly overlapped) subcells in $r$ by specifying different $j \in \{1, \dots, M\}$ as the indices, each of which satisfies $F_{(j)} \geq F_{(i)} \ \forall \ i \in \{1, \dots, M\}/j$.
More specifically, for any of such an index $j$, we assume that $\bm{x}$ is in an analytic cell $\tilde{r}_{(j)} \in \widetilde{\mathcal{R}}_{(j)}$, and override the notation $\tilde{r} = r = r_{(1)} \cap \cdots \cap \tilde{r}_{(j)} \cap \cdots \cap r_{(M)}$, which is the cell of interest relevant to extraction of zero-level isosurface --- we note that some of $r_{(i)}$, $i \in \{1, \dots, M\}/j$,
may also be analytic cells achieved by their respective subnetworks, and the subsequent analysis holds without explicit specification of their analytic cell status. Define $M-1$ inequalities $ (\bm{a}_{F_{(i)}}^{r_{(i)}} - \bm{a}_{F_{(j)}}^{\tilde{r}_{(j)}})\bm{x} \leq 0$, $i \in \{1, \dots, M\}/j$;
the kernels $\bm{a}_{F_{(i)}}^{r_{(i)}} - \bm{a}_{F_{(j)}}^{\tilde{r}_{(j)}}$ specify planes in $\mathbb{R}^3$ that partition $\tilde{r}$ as a convex polyhedral subcell, denoted as $\tilde{r}_{\textrm{\tiny Sub}}$. To specify the subcell, we compactly write the inequalities as
\begin{equation}\label{EqnAnalyticCellSystemMaxPoolingOperation}
  \left(\bm{A}^{/(j)} - \bm{1}_{M-1} \bm{a}_{F_{(j)}}^{\tilde{r}_{(j)}} \right) \bm{x} \preceq 0 ,
\end{equation}
where $\bm{A}^{/(j)} = [\bm{a}_{F_{(1)}}^{r_{(1)}}; \dots; \bm{a}_{F_{(j-1)}}^{r_{(j-1)}}; \bm{a}_{F_{(j+1)}}^{r_{(j+1)}}; \dots; \bm{a}_{F_{(M)}}^{r_{(M)}} ] \in \mathbb{R}^{(M-1) \times 3}$ and $\bm{1}_{M-1}$ is a vector with all its $M-1$ entries as the value $1$. Given that the cell $\tilde{r}$ can be explicitly determined by the set of boundary planes $\{ \bm{H}_{l{(i)}k{(i)}}^{r{(i)}}  \}$ with $\bm{H}_{l{(i)}k{(i)}}^{r{(i)}} = \{ \mathbf{x} \in \mathbb{R}^3 | \mathbf{a}^{r{(i)}}_{l{(i)}k{(i)}}\mathbf{x} = 0 \}$, where $l_{(i)} = 1, \dots, L_{(i)}$, $k_{(i)} = 1, \dots, n_{l_{(i)}}$, and $i \in \{1, \dots, M\}/j$, and the similarly defined set $\{ \bm{H}_{l{(j)}k{(j)}}^{\tilde{r}{(j)}}\}$,  the subcell of interest for the specified index $j$ can be written as
\begin{equation}\label{EqnAnalyticCellSystemMaxPooling}
  \mathbf{B}^{\tilde{r}_{\textrm{\tiny Sub}}}\mathbf{x}  =
  \begin{bmatrix}
    \bm{A}^{/(j)} - \bm{1}_{M-1} \bm{a}_{F_{(j)}}^{\tilde{r}_{(j)}}                              \\
    (\mathbf{I}_{(1)} - 2\textrm{diag}(\mathbf{s}(r_{(1)})) \mathbf{A}^{r_{(1)}}                 \\
    \vdots                                                                                       \\
    (\mathbf{I}_{(j)} - 2\textrm{diag}(\mathbf{s}(\tilde{r}_{(j)})) \mathbf{A}^{\tilde{r}_{(j)}} \\
    \vdots                                                                                       \\
    (\mathbf{I}_{(M)} - 2\textrm{diag}(\mathbf{s}(r_{(M)})) \mathbf{A}^{r_{(M)}}
  \end{bmatrix}
  \mathbf{x} \preceq 0 ,
\end{equation}
where $\mathbf{A}^{r_{(i)}}$, $i \in  \{1, \dots, M\}/j$, and $\mathbf{A}^{\tilde{r}_{(j)}}$ are defined similarly as (\ref{EqnAnalyticCellSystem}), and $\bm{I}_{(i)}$ is an identity matrix of compatible size. Given the system (\ref{EqnAnalyticCellSystemMaxPooling}), we have the analytic subcell associated with the union function $F$ defined as
\begin{equation}\label{EqnAnalyticCellMaxPooling}
  \mathcal{C}_F^{\tilde{r}_{\textrm{\tiny Sub}}} = \{ \mathbf{x} \in \mathbb{R}^3 | \mathbf{B}^{\tilde{r}_{\textrm{\tiny Sub}}}\mathbf{x} \preceq 0 \} .
\end{equation}
Since $\bm{a}_{F_{(j)}}^{\tilde{r}_{(j)}}\bm{x} \geq \bm{a}_{F_{(i)}}^{r_{(i)}}\bm{x} \ \forall \ i \in \{1, \dots, M\}/j$ in the present subcell, we have the corresponding analytic face defined as
\begin{equation}\label{EqnAnalyticFaceMaxPooling}
  \mathcal{P}_F^{\tilde{r}_{\textrm{\tiny Sub}}} = \{ \mathbf{x} \in \mathbb{R}^3 | \bm{a}_{F_{(j)}}^{\tilde{r}_{(j)}}\bm{x} = 0 , \mathbf{B}^{\tilde{r}_{\textrm{\tiny Sub}}}\mathbf{x} \preceq 0 \} .
\end{equation}
In practice, to implement the analytic marching algorithm proposed in Section \ref{SecAlgorithm}, for the present subcell $\tilde{r}_{\textrm{\tiny Sub}}$ specified by the index $j$, we define an additional state functional for the final max pooling operation of $F$ as
\begin{equation}\label{EqnStateFuntionalMaxPoolingOperation}
  \bm{s}_{\textrm{\tiny MaxPool}}(\bm{x}) = \bm{e}_j ,
\end{equation}
which is an $M$-dimensional one-hot vector with the entry of 1 at the $j^{th}$ index. Given the definitions (\ref{EqnAnalyticCellMaxPooling}), (\ref{EqnAnalyticFaceMaxPooling}), and (\ref{EqnStateFuntionalMaxPoolingOperation}), our theoretical analysis in Section \ref{SecTheory} and the analytic marching algorithm in Section \ref{SecAlgorithm} can be readily applied to such an architecture $F$ with a final max pooling aggregation. The state functional (\ref{EqnStateFuntionalMaxPoolingOperation}) is used, together with the state functionals of the $M$ subnetworks, to transit among analytic (sub-)cells during the analytic marching process.


\section{Learning Implicit Surface Networks for Generative Shape Modeling}
\label{SecApps}

The analysis and algorithm presented in the previous sections assume that an implicit function $F = f\circ \bm{T}$ has been given. In this section, we present different manners to construct and learn $F$ for shape modeling and reconstruction. These manners have their respective advantages when $F$ represents an implicit field of SDF or an occupancy field.

\subsection{Learning as a Direct Shape Encoding}
\label{SecDirectShapeEncoding}

We first show the usefulness of analytic marching by directly fit individual 3D shapes to a model $F = f\circ \bm{T}$ constructed from a ReLU based MLP. A similar strategy is taken in \cite{davies2021effectiveness} that demonstrates the compactness of neural network as a shape representation. Take a field of SDF as the example.
Assume that a surface $\mathcal{M}$ to be encoded is given. Following \cite{Gropp2020}, we train the network with a regularized objective
\begin{equation}\label{EqnDirectEncodingObj}
  \min_{F=f \circ \boldsymbol{T}} \ell_{\mathcal{M}}(F) + \lambda_1^{\textrm{\tiny Direct}} \mathbb{E}_{\boldsymbol{x}\sim \mathbb{R}^3} \bigl\lvert \Vert \nabla_{\boldsymbol{x}} F(\boldsymbol{x}) \Vert_2 - 1 \bigr\rvert
  ,
\end{equation}
with
\begin{equation}\label{EqnDirectEncodingManifoldLoss}
  \ell_{\mathcal{M}}(F) = \mathbb{E}_{\bm{z}\sim \mathcal{M}} \left[ |F(\bm{z})| + \lambda_2^{\textrm{\tiny Direct}} \Vert \nabla_{\bm{z}} F(\bm{z}) - \bm{n}_{\bm{z}} \Vert_2 \right] ,
\end{equation}
where $\lambda_1^{\textrm{\tiny Direct}}$ and $\lambda_2^{\textrm{\tiny Direct}}$ are penalty parameters, and $\bm{n}_{\bm{z}}$ denotes the normal vector at a surface point $\bm{z} \in \mathcal{M}$; the second term of (\ref{EqnDirectEncodingManifoldLoss}) is optionally used when surface normals are available \cite{Gropp2020}. 

To improve the training efficiency, one may also replace the ReLU nonlinearity with
a smooth version $g_{\alpha}(x) = x e^{\alpha x} / (1 + e^{\alpha x})$ \cite{Lange2014} during training,
where $\alpha > 0$ is a parameter controlling the degree of approximation, which is gradually increased until the training convergence. Note that after training, we still use the standard ReLU nonlinearity for analytic marching.

%

\subsection{Global Decoding for Reconstruction of Novel Surface Shapes}
\label{SecGlobalDecoding4Novel}

Learning deep models to reconstruct novel shape instances gains popularity in recent research of deep learning surface reconstruction \cite{Park2019,Mescheder2019}. Given training shapes, these methods usually train an encoder-decoder architecture for the purpose. Considering that the training shapes are point clouds sampled from ground-truth object surfaces, these methods train a point set encoder (e.g., a PointNet \cite{Charles2017}) that outputs latent shape representation for a testing point cloud, which, together with sampled points in the 3D implicit space, are then fed into the decoder for inference of the implicit surface. Heavy MLP decoders are usually used in order for learning to generalize to novel shape instances. Given that the computation complexity of our analytic marching depends on the network capacities (cf. Section \ref{SecComplexityAnalysis}), we choose to use a hypernetwork \cite{Littwin2019} for shape decoding, instead of directly using an MLP decoder. More specifically, the hypernetwork can be chosen as a heavy MLP, which takes as input a latent shape representation from the encoder, and outputs weights of another light MLP, and the resulting light MLP is used as the implicit model $F = f\circ\bm{T}$ for surface inference via analytic marching. Such a hypernetwork based pipeline enjoys the benefit of precisely modeling and decoding novel shapes, while keeping an efficient process of shape inference.

Let $E$ and $H$ respectively denote the encoder and hypernetwork. Given a training set of ground-truth surfaces $\{ \mathcal{M} \in \mathfrak{M} \}$, each of which is sampled from the distribution $\mathfrak{M}$, we use the following objective to train $E$ and $H$
\begin{equation}\label{EqnGlobalDecodingObj}
  \min_{E, H} \mathbb{E}_{\mathcal{M}\sim \mathfrak{M}} \ell_{\mathbb{R}^3}(E,H;\mathcal{M}) + \lambda^{\textrm{\tiny Global}} \Vert E(\mathcal{M})\Vert_2^2 ,
\end{equation}
with
\begin{equation}
  \ell_{\mathbb{R}^3}(E,H;\mathcal{M}) = \mathbb{E}_{\boldsymbol{x}\sim \mathbb{R}^3} |F(\bm{x}; H(E(\mathcal{M}))) - d(\boldsymbol{x}; \mathcal{M})| , \nonumber
\end{equation}
where $H(E(\mathcal{M}))$ outputs network weights of an implicit $F(\cdot; H(E(\cdot)))$, and $d(\boldsymbol{x}; \mathcal{M})$ is the ground-truth signed distance of $\bm{x} \in \mathbb{R}^3$ to the surface $\mathcal{M}$; the latent representation $E(\mathcal{M})$ in (\ref{EqnGlobalDecodingObj}) is regularized with its $L_2$ norm, and $\lambda^{\textrm{\tiny Global}}$ is a penalty parameter.

\subsection{Improved Reconstruction via an Ensemble of Local Decoders}
\label{SecEnsembleOfLocalDecoders}

Global decoding either via a direct MLP or indirectly via a hypernetwork is limited in reconstructing surface shapes of complex topologies \cite{Chen2019,Tretschk2020}. To remedy, one strategy is to rely on local models and reconstruct a topologically complex surface as a union of local surface parts, i.e., a typical strategy in constructive solid geometry \cite{Botsch2010}.

In this work, we technically implement this strategy using an ensemble of local decoders, as illustrated in Fig. \ref{FigExtOtherArchitecture}-(c).
Let $E$ be the shape encoder, and $F_{(i)}$, $i = 1, \dots, M$, denote the local decoders. Given the latent representation $E(\mathcal{M})$ for a surface $\mathcal{M}$, we again use a hypernetwork $H$ to estimate weights of $\{ F_{(i)} \}_{i=1}^M$. As indicated by (\ref{EqnMaxPoolingForwardPropagation}), the global implicit function $F$ can be constructed as a maximum over outputs of the local decoders, i.e., $F(\bm{x}) = \max_{i \in \{1, \ldots, M\}}  F_{(i)} (\bm{x})$ for $\bm{x} \in \mathbb{R}^3$. Indeed, constructive solid geometry suggests that any boolean operation of occupancy fields can be realized by maximum function, and therefore composition of several components is the maximal value of individual occupancy values.
The max function used in (\ref{EqnMaxPoolingForwardPropagation}) can also be replaced as a soft version \cite{Lange2014} to improve the training efficiency, i.e.,
\begin{equation}
  F(\bm{x}) =
  \frac{\sum_{i=1}^M F_{(i)}(\bm{x}) e^{\beta F_{(i)}(\bm{x})}}{\sum_{i=1}^M e^{\beta F_{(i)}(\bm{x})}},
\end{equation}
where $\beta > 0$ is a parameter controlling the degree of softness, which is gradually increased during training. Given a training set of ground-truth surfaces $\{ \mathcal{M} \in \mathfrak{M} \}$, we use the following regularized objective to train such a model on occupancy fields
\begin{equation}\label{EqnLocalDecodingObj}
  \min_{E, H} \mathbb{E}_{\mathcal{M}\sim \mathfrak{M}} \ell_{\mathbb{R}^3}(E,H;\mathcal{M}) + \lambda_1^{\textrm{\tiny Local}} \Vert E(\mathcal{M})\Vert_2^2 + \lambda_2^{\textrm{\tiny Local}} \Vert H(E(\mathcal{M}))\Vert_2^2 ,
\end{equation}
with
\begin{equation}
  \ell_{\mathbb{R}^3}(E,H;\mathcal{M}) =
  \mathbb{E}_{\boldsymbol{x}\sim \mathbb{R}^3} \textrm{CE} \left( F(\boldsymbol{x}; H(E(\mathcal{M}))), o(\boldsymbol{x}; \mathcal{M})\right) , \nonumber
\end{equation}
where $\textrm{CE}(\cdot, \cdot)$ denotes a binary cross-entropy loss, $o(\boldsymbol{x}; \mathcal{M}) \in \{0, 1\}$ indicates the occupancy status of $\bm{x} \in \mathbb{R}^3$, and $\lambda_1^{\textrm{\tiny Local}}$ and $\lambda_2^{\textrm{\tiny Local}}$ are penalty parameters. In (\ref{EqnLocalDecodingObj}), we also use an $L_2$-norm regularization to constrain weights of the local decoders estimated from $H$, which is effective to regularize the learning.


\section{Experiments}
\label{SecExp}

\noindent\textbf{Datasets}
We use two datasets of 3D solid objects for our experiments. The first dataset consists of object instances from five categories of the ShapeNet \cite{ShapeNet} (namely, ``Rifle'', ``Chair'', ``Airplane'', ``Sofa'', and ``Table''). We construct our second, Richly Detailed (RD), dataset by collecting 5 geometrically and topologically complex shapes from Stanford 3D Scanning Repository \cite{Stanford3DScanningRepository}, Artec3D \cite{Artec3D}, and Free3D \cite{Free3D}; Fig. \ref{FigExpsDirectShapeEncodingVisualizations} and Appendix \ref{AppendixRDAdditionalResults} show these shapes. We normalize object mesh models of the two datasets in the unit sphere of the 3D space. We use both the two datasets for our experiments of direct shape encoding (cf. Section \ref{SecDirectShapeEncoding}). For experiments of learning to reconstruct novel object shapes (cf. Sections \ref{SecGlobalDecoding4Novel} and \ref{SecEnsembleOfLocalDecoders}), we use the first dataset from ShapeNet where we split object instances of each category as training or test ones by a ratio of 4:1, and obtain input point clouds for the encoder $E(\cdot)$ by sampling 2,048 points from each ground-truth mesh; when implementing the training objectives (\ref{EqnGlobalDecodingObj}) and (\ref{EqnLocalDecodingObj}), ground-truth values of SDF or occupancy are calculated by linear interpolation from a dense grid obtained by \cite{Xu2014, Sin2013}.

\vspace{0.1cm}
\noindent\textbf{Implementation Details}
For direct shape encoding, we use an MLP of width 60 and depth 8. The network is trained for $1,500$ epochs, with learning rates starting from $1 \times 10^{-3}$ and dropping by a factor of $0.3$ at epochs $1100$, $1200$, $1350$ and $1450$; we set $\lambda_1^{\textrm{\tiny Direct}} = 0.2$, $\lambda_2^{\textrm{\tiny Direct}} = 1.0$, and the slope $\alpha$ used in soft ReLU is initialized to $10$ and gradually increased to $10,000$.
For experiments of learning to reconstruct novel shapes, we directly use PointNet \cite{Charles2017} as the encoder $E(\cdot)$.
We use a hypernetwork of MLP with depth 2 and width 1,024 for global decoding, which gives a decoding MLP of width 60 and depth 6; we set $\lambda^{\textrm{\tiny Global}} = 0.01$; learning rates start from $3 \times 10^{-4}$ and drop at the epoch $2,400$, $4,200$, and $5,400$ respectively by a factor of $0.2$, until a total of $6,000$ epochs.
For local decoding, we again use a hypernetwork of MLP with depth 2 and width 1,024, which results in an ensemble of $M = 4$ local decoders; each local decoder is an MLP of width 32 and depth 4; we set $\lambda_1^{\textrm{\tiny Local}} = 0.01$ and $\lambda_2^{\textrm{\tiny Local}} = 5 \times 10^{-5} $; the hypernetwork is trained for $3,000$ epochs, and learning rates start from $3 \times 10^{-4}$ and are reduced by a factor of $0.2$ at the epoch $1,200$, $2,100$, and $2,700$; during inference, individual shape components are independently extracted and then integrated using \cite{Douze2017}.
All experiments are conducted on a single Nvidia Tesla K80.

\vspace{0.1cm}
\noindent\textbf{Comparative Methods and Evaluation Metrics}
We compare our proposed meshing algorithm of Analytic Marching (AM) with existing ones, which have already been the standard meshing choices given that implicit functions are provided; the comparative methods include Greedy Meshing (GM), Marching Cubes (MC) \cite{Lorensen1987}, Marching Tetrahedra (MT) \cite{Doi1991}, and Dual Contouring (DC) \cite{Ju2002}. All these methods are based on discrete sampling of the 3D space for evaluation of signed distances or occupancies for the sampled points, which are then used for extraction of the zero-level isosurfaces. They are thus by nature different from our AM. \emph{We emphasize that the comparisons are made on meshing algorithms themselves, independent of how the implicit functions have been constructed or learned; our presented methods of learning implicit surface networks in Section \ref{SecApps} are mainly to set contexts for such a meshing comparison.} As such, we use evaluation metrics of (approximate) distances between each ground-truth mesh and those extracted by different meshing algorithms, including Chamfer Distance (CD) and Earch Mover Distance (EMD), each of which computes symmetric, pairwise Euclidean distance between sampled point sets, Intersection over Union (IoU) that measures how the two meshes overlap and whose values range in [0, 1], and F-score (F) that measures the symmetrical percentage of reachable surface areas within a given distance $\tau$; we use a default $\tau = 5\times10^{-3}$. In addition, we report relevant attributes of meshing algorithms and results, such as the number of triangular faces per mesh (\textsharp TriFace) and running time, where triangular faces of the results from our AM are converted from polygonal ones (cf. Section \ref{SecMeshSimplification} for how the conversion can be conducted).


\subsection{Analysis of Analytic Marching}
\label{SecExpsAnalysisAM}

In this section, we analyze various properties of analytic marching. Experiments are conducted on instances of ShapeNet by directly fitting implicit MLP functions.

\vspace{0.1cm}
\noindent\textbf{Effects of Network Capacities}
Section \ref{SecBasicAM} suggests that the meshing precision of our AM depends on the (maximal) number of linear regions partitioned by an MLP, whose order is exponential to network depth and polynomial to network width. To verify empirically, we design experiments by using two groups of MLPs that respectively have the same numbers of 360 and 900 neurons. The first group distributes their neurons as D4-W90, D6-W60, and D8-W45, where ``D'' is for depth and ``W'' is for width, and the second group distributes their neurons as D10-W90, D15-W60, and D20-W45. Results in Table \ref{TableExpsArchitectures} confirm that mesh accuracies increase consistently with the increased network capacities, but at a cost of much increased face numbers per mesh. Given the same number of neurons, it seems that a balanced depth-width neuron distribution is more advantageous at the studied regime of relatively lower network capacities.

\begin{table}[!htbp]
  \caption{Meshing accuracies of analytic marching by using MLPs of different capacities. ``D'' stands for network depth and ``W'' stands for network width. Results are obtained by averaging over 200 instances of the ``Rifle'' category from ShapeNet.  } \vspace{-0.8cm}
  \label{TableExpsArchitectures}
  \vskip 0.1in
  \begin{center}
    \begin{small}
      \begin{tabular}{|c|c|c|c|c|c|}
        \hline
        Architecture & CD $\downarrow$   & EMD $\downarrow$    & IoU $\uparrow$  & F@$\tau$ $\uparrow$ & \textsharp TriFace \\
        \hline
        \hline
        D4-W90       & 0.616 & 0.00860 & 0.865 & 0.848       & 195,658            \\
        \hline
        D6-W60       & 0.529 & 0.00800 & 0.869 & 0.855       & 200,263            \\
        \hline
        D8-W45       & 0.567 & 0.00812 & 0.857 & 0.842       & 186,903            \\
        \hline
        \hline
        D10-W90      & 0.364 & 0.00585 & 0.902 & 0.881       & 844,114            \\
        \hline
        D15-W60      & 0.385 & 0.00623 & 0.889 & 0.874       & 671,752            \\
        \hline
        D20-W45      & 0.421 & 0.00689 & 0.861 & 0.866       & 507,742            \\
        \hline
      \end{tabular}
    \end{small}
  \end{center}
  \vskip -0.1in
\end{table}

\vspace{0.1cm}
\noindent\textbf{Results of Different Categories}
To investigate how AM performs on instances of different categories whose surface complexities may vary, we conduct experiments on five categories of ShapeNet, using the MLP of D6-W60 as mentioned above. Table \ref{TableExpsComparisons} shows that the category of ``Airplane'' has the best meshing accuracies in terms of 3 of the total 4 metrics, which is in accordance with the common intuition on the simplicity of its shapes.

\begin{table}[!htbp]
  \caption{Meshing accuracites of analytic marching on five categories of ShapeNet, using an MLP of depth 6 and width 60 (the same D6-W60 network as in Table \ref{TableExpsArchitectures}). Results are obtained by averaging over 200 instances of each category.  } \vspace{-0.6cm}
  \label{TableExpsComparisons}
  \vskip 0.1in
  \begin{center}
    \begin{small}
      \begin{tabular}{|c|c|c|c|c|c|}
        \hline
        Category & CD $\downarrow$   & EMD  $\downarrow$   & IoU $\uparrow$  & F@$\tau$ $\uparrow$  & \textsharp TriFace \\
        \hline
        \hline
        Rifle    & 0.529 & 0.00800 & 0.869 & 0.855         & 200,263            \\
        \hline
        Chair    & 0.727 & 0.00801 & 0.896 & 0.529         & 365,231            \\
        \hline
        Airplane & 0.325 & 0.00532 & 0.894 & 0.898         & 262,116            \\
        \hline
        Sofa     & 0.678 & 0.00665 & 0.966 & 0.532         & 313,718            \\
        \hline
        Table    & 0.698 & 0.00716 & 0.907 & 0.516         & 326,808            \\
        \hline
      \end{tabular}
    \end{small}
  \end{center}
  \vskip -0.1in
\end{table}

\vspace{0.1cm}
\noindent\textbf{Efficiencies of Customized Triggering Schemes}
We present different triggering schemes of AM in Section \ref{SecCustomizedInitialization} as improvements over the simple one of solving (\ref{EqnSurfacePointOptim}) via SGD, including Sphere Tracing (ST) and Dichotomy (DICH). To verify their efficiencies, we conduct experiments of finding 1,024 points on the surface for each instance of all the five categories of ShapeNet, using the aforementioned MLP of D6-W60. We compare the three schemes in terms of their required numbers of iterations (\textsharp Iter) to converge, the corresponding running time (second), and the rate of success (SR). Table \ref{TableExpsTriggering} tells that for signed distance field (SDF), all the schemes can successfully converge to find the points on the surface and thus trigger the AM algorithm, and ST and DICH are much faster than SGD; for occupancy field (OF), DICH is the only scheme that can trigger AM successfully and efficiently. We thus recommend DICH as the default triggering scheme of AM.

\begin{table*}[!t]
  \caption{Comparisons among different triggering scheme of analytic marching (cf. Section \ref{SecCustomizedInitialization}). Results are obtained by finding 1,024 points on the surface for each instance of all the five categories of ShapeNet, using an MLP of depth 6 and width 60. We use measures of the required number of iterations (\#Iter) to converge, the corresponding running time (second), and the success rate (SR). The mark ``-'' indicates the failure of triggering. } \vspace{-0.2cm}
  \label{TableExpsTriggering}
  \vspace{-0.3cm}
  \begin{center}
    \scalebox{1.0}{
      \begin{tabular}{cccccccccc}
        \hline
        \multicolumn{1}{|c|}{Method}        & \multicolumn{3}{c|}{SGD}                                                                                                & \multicolumn{3}{c|}{ST}                                                                                                   & \multicolumn{3}{c|}{DICH}                                                                                                                                                                                                                                                   \\
        \hline
        \multicolumn{1}{|C{1.3cm}|}{Metric} & \multicolumn{1}{C{1.3cm}|}{\textsharp Iter} & \multicolumn{1}{C{1.3cm}|}{Time(sec.)} & \multicolumn{1}{C{1.3cm}|}{SR}  & \multicolumn{1}{C{1.3cm}|}{\textsharp Iter} & \multicolumn{1}{C{1.3cm}|}{Time(sec.)}   & \multicolumn{1}{C{1.3cm}|}{SR}  & \multicolumn{1}{C{1.3cm}|}{\textsharp Iter} & \multicolumn{1}{C{1.3cm}|}{Time(sec.)}   & \multicolumn{1}{C{1.3cm}|}{SR}  \\
        \hline
        \hline
        \multicolumn{1}{|C{1.3cm}|}{SDF}     & \multicolumn{1}{C{1.3cm}|}{531}              & \multicolumn{1}{C{1.3cm}|}{3.47}       & \multicolumn{1}{C{1.3cm}|}{1.0} & \multicolumn{1}{C{1.3cm}|}{3}                & \multicolumn{1}{C{1.3cm}|}{0.0267}       & \multicolumn{1}{C{1.3cm}|}{1.0} & \multicolumn{1}{C{1.3cm}|}{11}               & \multicolumn{1}{C{1.3cm}|}{0.0657}       & \multicolumn{1}{C{1.3cm}|}{1.0} \\
        \hline
        \multicolumn{1}{|C{1.3cm}|}{OF}      & \multicolumn{1}{C{1.3cm}|}{-----}            & \multicolumn{1}{C{1.3cm}|}{-----}      & \multicolumn{1}{C{1.3cm}|}{0.0} & \multicolumn{1}{C{1.3cm}|}{-----}            & \multicolumn{1}{C{1.3cm}|}{-----}        & \multicolumn{1}{C{1.3cm}|}{0.0} & \multicolumn{1}{C{1.3cm}|}{22}               & \multicolumn{1}{C{1.3cm}|}{0.0933}       & \multicolumn{1}{C{1.3cm}|}{1.0} \\
        \hline
      \end{tabular}
    }
  \end{center}
\end{table*}

\vspace{0.1cm}
\noindent\textbf{Parallel Marching with CUDA Implementation} Our AM algorithm supports parallel marching, which simultaneously marches the analtyic cells to recover the mesh. In this work, we implement parallel marching with CUDA implementation. Table \ref{TableExpsCUDAvsCPU} shows that on Nvidia GPUs (64-bit floating point), parallel marching with the ST triggering of 1,024 points on the surface is nearly an order-of-magnitude faster than solving AM on CPUs using SGD triggering of a same number of points. These experiments are conducted on all the five categories of ShapeNet, using the MLP of D6-W60 as mentioned above.

\begin{table}[!htbp]
  \caption{Efficiency of parallel marching with CUDA implementation. Experiments are conducted on a Nvidia Tesla K80 (64-bit floating point) using an MLP of depth 6 and width 60 for SDF modeling. Results are averaged over instances of all the five categories of ShapeNet. SGD, ST, and DICH are the three triggering schemes discussed in Section \ref{SecCustomizedInitialization}. } \vspace{-0.6cm}
  \label{TableExpsCUDAvsCPU}
  \vskip 0.1in
  \begin{center}
    \begin{small}
      \scalebox{1.0}{
        \begin{tabular}{|c|c|}
          \hline
          Implementation of Parallel Marching & Time (second) \\
          \hline
          \hline
          CPU + SGD       & 20.8       \\
          \hline
          CUDA + SGD      & 6.22       \\
          \hline
          CUDA + DICH       & 2.82      \\
          \hline
           CUDA + ST       & 2.78       \\
          \hline
        \end{tabular}
      }
    \end{small}
  \end{center}
  \vskip -0.1in
\end{table}

\begin{figure*}[!htbp]
  \vskip 0.1in
  \begin{center}
    \includegraphics[scale=0.12]{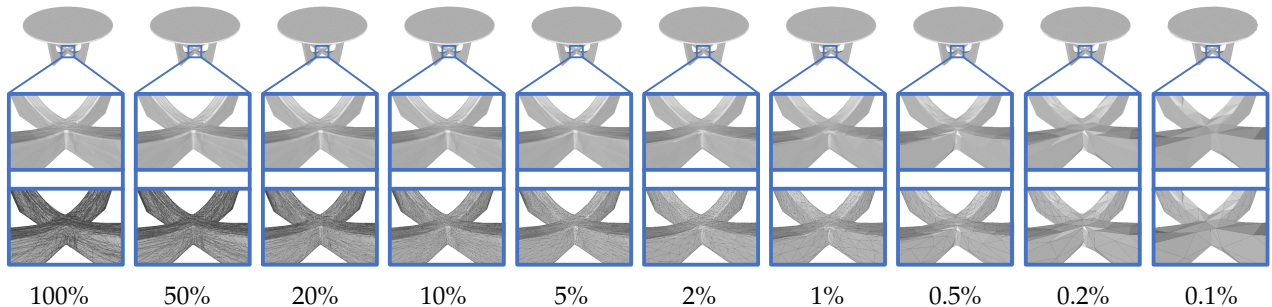} 

    \scalebox{1.0}
    {
      \begin{tabular}{C{1.28cm} C{1.28cm} C{1.28cm} C{1.28cm} C{1.28cm} C{1.28cm} C{1.28cm} C{1.28cm} C{1.28cm} C{1.28cm}}
         100\% &  50\% &  20\% &  10\% & 5\% &  2\% & 1\% & 0.5\% & 0.2\% & 0.1\% \\
      \end{tabular}
    }
    \vspace{-0.3cm}
    \caption{Example results of analytic marching after mesh simplification at different ratios. Better viewing by zooming in the electronic version.      }
    \label{FigExpsSimplificationVisualizations}
  \end{center}
  \vskip -0.1in
\end{figure*}

\vspace{0.1cm}
\noindent\textbf{Mesh Simplification in \texttt{AnalyticMesh}}
As indicated by the results in Tables \ref{TableExpsArchitectures} and \ref{TableExpsComparisons}, in spite of the exactness, AM tends to produce polygon meshes with huge numbers of polygon faces. This brings inconvenience to downstream processing on meshes. It is desirable to simplify the meshes at no or less sacrifice of mesh precisions. As stated in Section \ref{SecMeshSimplification}, we have incorporated such postprocessing operations into our publicly released package of \texttt{AnalyticMesh}. Quantitative results in Table \ref{TableExpsSimplificationNumerics} show that by mesh simplification, the loss of precision is negligible even when preserving only 1\% of the original numbers of mesh faces. A corresponding visualization is shown in Fig. \ref{FigExpsSimplificationVisualizations}. \emph{Except mentioning otherwise, we present all results of AM in the subsequent sections after mesh simplification at a ratio of 10\%.}

\begin{table}[!htbp]
  \caption{Quantitative results of mesh simplification in \texttt{AnalyticMesh}. Mesh results are obtained by first fitting an MLP of depth 6 and width 60 to each instance of the five categories of ShapeNet, and then recovering the mesh via AM and simplifying the mesh using the corresponding operations in \texttt{AnalyticMesh}. Note that the simplification is controlled by quadric error metrics \cite{Garland1997}, and a specified ratio does not translate exactly as a correspondingly reduced number of mesh faces. } \vspace{-0.6cm}
  \label{TableExpsSimplificationNumerics}
  \vskip 0.1in
  \begin{center}
    \begin{small}
      \scalebox{0.85}{
        \begin{tabular}{|c|c|c|c|c|c|c|c|}
          \hline
          Ratio & CD $\downarrow$      & EMD $\downarrow$  & IoU $\uparrow$  & F@$\tau$ $\uparrow$  & \textsharp TriFace & Mesh Size   \\
          \hline
          \hline
          100\%     & 0.591            & 0.00703           & 0.906           & 0.666                & 293,627            & 4,161 KB   \\
          \hline
          50\%      & 0.591            & 0.00704           & 0.906           & 0.666                & 148,686            & 2,093 KB    \\
          \hline
          20\%      & 0.591            & 0.00704           & 0.906           & 0.666                & 59,218             & 835 KB   \\
          \hline
          10\%      & 0.591            & 0.00705           & 0.905           & 0.666                & 29,342             & 417 KB    \\
          \hline
          5\%       & 0.592            & 0.00713           & 0.905           & 0.665                & 14,762             & 209 KB   \\
          \hline
          2\%       & 0.594            & 0.00722           & 0.903           & 0.665                & 5,909              & 83.7 KB  \\
          \hline
          1\%       & 0.599            & 0.00730           & 0.901           & 0.664                & 2,964              & 42.0 KB   \\
          \hline
          0.5\%     & 0.608            & 0.00768           & 0.899           & 0.663                & 1,480              & 21.1 KB   \\
          \hline
          0.2\%     & 0.797            & 0.00996           & 0.882           & 0.653                & 586                & 8.54 KB    \\
          \hline
          0.1\%     & 1.59             & 0.0134            & 0.853           & 0.634                & 294                & 4.37 KB   \\
          \hline
        \end{tabular}
      }
    \end{small}
  \end{center}
  \vskip -0.1in
\end{table}

\subsection{Direct Shape Encoding}
\label{ExpDirectShapeCoding}

In this section, we compare our proposed AM with existing methods in the context of direct shape encoding, where an implicit network of SDF is trained to fit each shape instance (cf. Section \ref{SecDirectShapeEncoding} for the details). Experiments are conducted on both the five categories of ShapeNet and the RD dataset. For ShapeNet, we use an MLP of depth 6 and width 60 (the same D6-W60 model as in the preceding section); for RD, we use an MLP of depth 8 and width 60.

\begin{figure*}[!htbp]
  \vskip 0.1in
  \begin{center}
    \centerline{\includegraphics[scale=0.242]{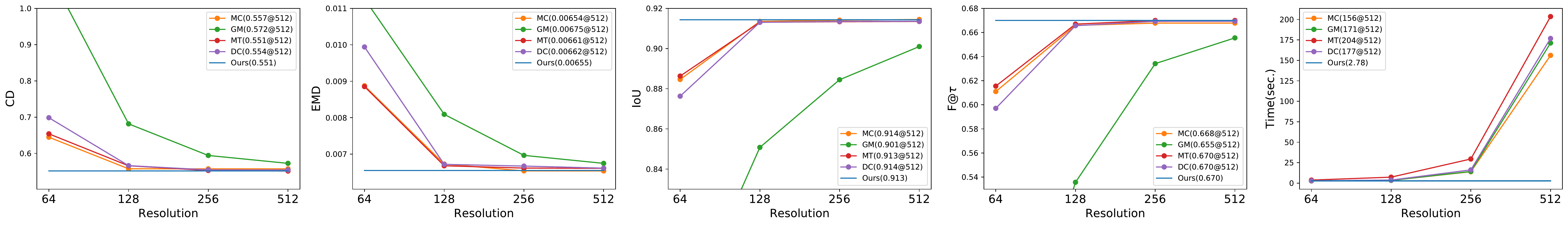}}  \vspace{-0.4cm}
    \caption{Quantitative comparisons of direct shape encoding under metrics of recovery precision and inference time. For greedy meshing (GM), marching cubes (MC), marching tetrahedra (MT), and dual contouring (DC), results under a resolution range of discrete point sampling from $64^3$ to a GPU memory limit of $512^3$ are presented. Experiments are conducted on shape instances of five categories from ShapeNet, using an MLP of depth 6 and width 60 for SDF modeling.   }
    \label{FigExpsComparisons}
  \end{center}
  \vskip -0.1in
\end{figure*}

The plotting of numerical results on ShapeNet instances is shown in Fig. \ref{FigExpsComparisons}. Under different evaluation metrics, mesh accuracies of existing methods are upper bounded by our proposed AM. These methods of marching cubes family recover a mesh by firstly sampling a grid of 3D points at a specified resolution, followed by identifying the intersections with the zero-level isosurface; as such, their accuracies depend on the sampling resolutions. Instead, our AM recovers the mesh exactly captured by the fitted MLP.

\begin{figure*}[!htbp]
  \vskip 0.1in
  \begin{center}
    \includegraphics[scale=0.289]{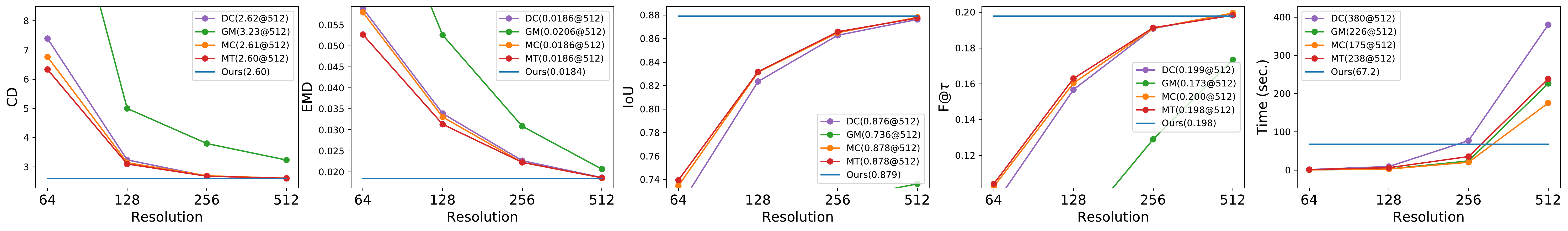} \vspace{-0.4cm}
    \caption{
      Quantitative comparisons of direct shape encoding under metrics of recovery precision and inference time. For greedy meshing (GM), marching cubes (MC), marching tetrahedra (MT), and dual contouring (DC), results under
      a resolution range of discrete point sampling from $64^3$ to a GPU memory limit of $512^3$ are presented. Experiments are conducted on five geometrically and topologically complex shape instances, using an MLP of depth 8 and width 60 for SDF modeling.
    }
    \label{FigExpsDirectShapeEncodingNumerics}
  \end{center}
  \vskip -0.1in
\end{figure*}

\begin{figure*}[!htbp]
    \vskip 0.1in
    \begin{center}

      \hspace{0.0cm}\includegraphics[scale=0.084]{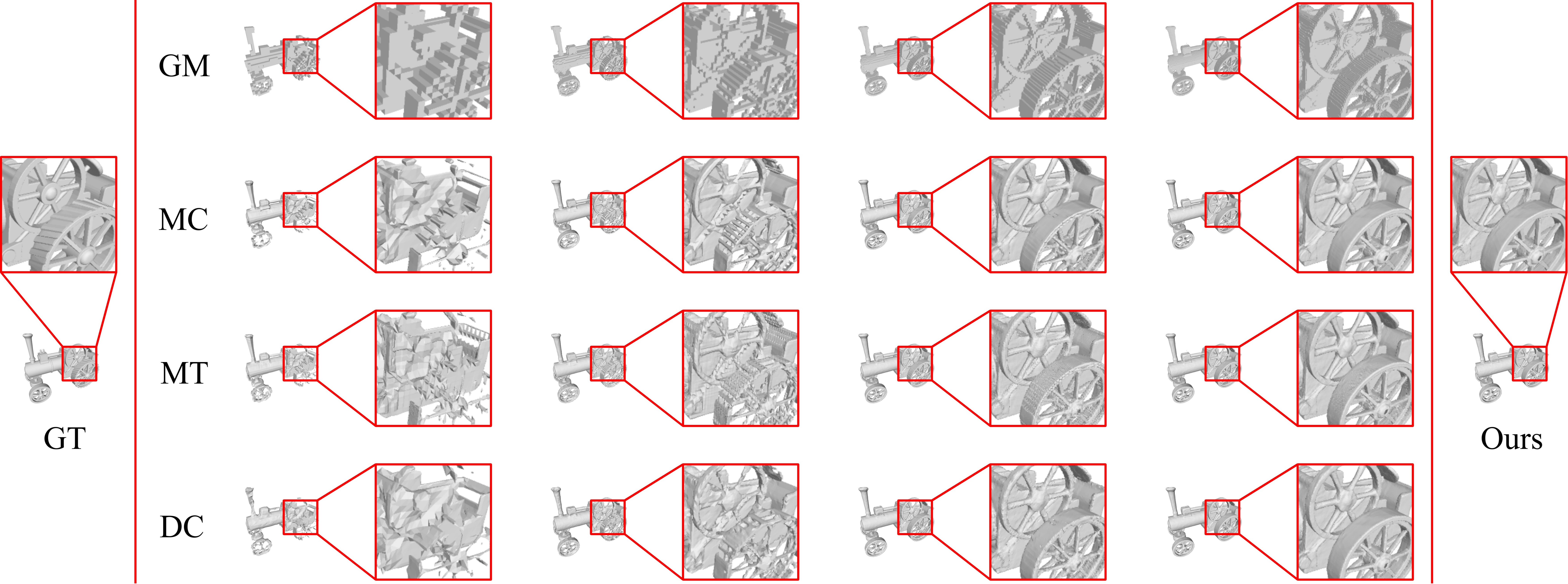}
      \vspace{0.2cm}

      \hspace{0.0cm}\includegraphics[scale=0.084]{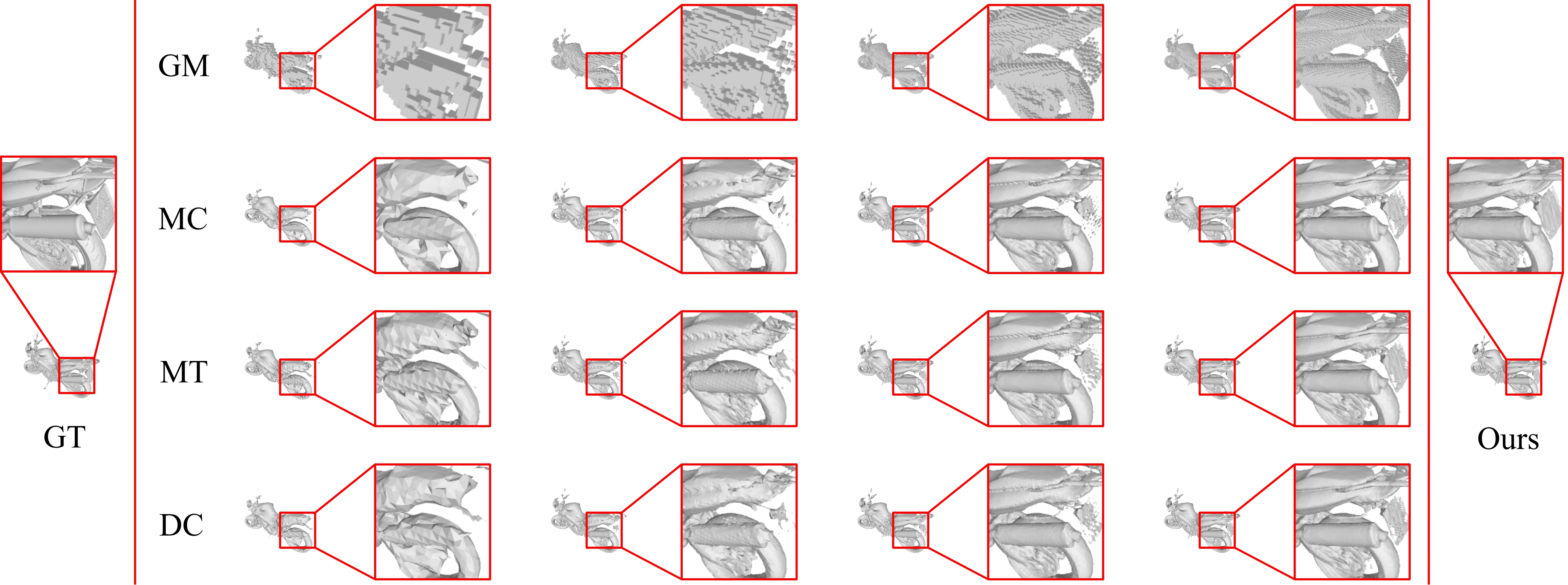}
      \vspace{0.02cm}

      \hspace{1.5cm}\scalebox{1.0}
      {
        \begin{tabular}{C{3.2cm} C{3.2cm} C{3.2cm} C{3.2cm}}
           MC64 &  MC128 &  MC256 & MC512
        \end{tabular}
      }

      \vspace{-0.1cm}
      \caption{Qualitative comparisons of direct shape encoding. The shown shape instance is fitted to an MLP of depth 8 and width 60 for SDF modeling. For greedy meshing (GM), marching cubes (MC), marching tetrahedra (MT), and dual contouring (DC), results under a resolution range of discrete point sampling from $64^3$ to a GPU memory limit of $512^3$ are presented. Better viewing by zooming in the electronic version. }
      \label{FigExpsDirectShapeEncodingVisualizations}
    \end{center}

    \vskip -0.1in
  \end{figure*}

Fig. \ref{FigExpsDirectShapeEncodingNumerics} shows the plotting of numerical results on the five shape instances of RD dataset. Example results for one of the shapes are given in Fig. \ref{FigExpsDirectShapeEncodingVisualizations}. Mesh accuracies of existing methods are still upper bounded by our proposed AM. We also notice from Fig. \ref{FigExpsDirectShapeEncodingVisualizations} that existing methods may give geometrically and/or topologically wrong recoveries, even when their sampling resolutions are increased to $512^3$. In contrast, AM gives much better and visually pleasant results. Results of other shape instances from the RD dataset are given in Appendix \ref{AppendixRDAdditionalResults}.

\subsection{Global Learning for Novel Shape Reconstruction}
\label{ExpGlobalLearning}

In this section, we compare our proposed AM with existing meshing methods in the context of learning a global model for decoding of novel shape instances. We train a hypernetwork of MLP with depth 2 and width 1,024, which learns weights of a light MLP with depth 6 and width 60 for SDF modeling. Experiments are conducted on the five ShapeNet categories by training the hypernetwork on the training shapes and testing on novel ones. Other learning setups are given in the beginning of Section \ref{SecExp}. Quantitative results in Table \ref{TableExpsGlobalShapeLearning} show that AM, \emph{after mesh simplification at a ratio of 10\%}, achieves mesh reconstruction accuracies nearly identical to those of MC512 (i.e., marching cubes with sampling of $512^3$ 3D points), while running much faster with much fewer numbers of mesh faces; this is due to the better accuracies of the original meshes exactly recovered by AM. The qualitative comparisons given in Appendix \ref{AppendixNovelShapeReconComp} are consistent to those quantitative ones.

\begin{table*}[!t]
  \caption{Quantitative comparisons between analytic marching (AM) and marching cubes (MC) in the context of learning a global model for decoding of novel shape instances. Experiments are conducted on shape instances of five categories from ShapeNet, by training a hypernetwork that gives weights of a light MLP with depth 6 and width 60 for SDF modeling. \emph{Results of AM are after simplification of the originally recovered meshes at a ratio of 10\%}. For MC, the sampling resolutions of 3D points range from $64^3$ to $512^3$.  } \vspace{-0.2cm}
  \label{TableExpsGlobalShapeLearning}
  \vspace{-0.3cm}
  \begin{center}
    \scalebox{0.8}{
      \begin{tabular}{cccccccccccccccc}
        \hline
        \multicolumn{1}{|C{1.20cm}|}{Metric}   & \multicolumn{5}{c|}{CD}                                                                                                                                                                                                & \multicolumn{5}{c|}{\textsharp TriFace}                                                                                                                                                                    & \multicolumn{5}{c|}{Time (sec.)}                                                                                                                                                                     \\
        \cline{1-16}
        \multicolumn{1}{|C{1.20cm}|}{Method}   & \multicolumn{1}{C{0.90cm}|}{MC64}       & \multicolumn{1}{C{0.90cm}|}{MC128} & \multicolumn{1}{C{0.90cm}|}{MC256}          & \multicolumn{1}{C{0.90cm}|}{MC512}          & \multicolumn{1}{C{0.90cm}|}{AM}             & \multicolumn{1}{C{0.90cm}|}{MC64}           & \multicolumn{1}{C{0.90cm}|}{MC128} & \multicolumn{1}{C{0.90cm}|}{MC256}  & \multicolumn{1}{C{0.90cm}|}{MC512}  & \multicolumn{1}{C{0.90cm}|}{AM}             & \multicolumn{1}{C{0.90cm}|}{MC64} & \multicolumn{1}{C{0.90cm}|}{MC128} & \multicolumn{1}{C{0.90cm}|}{MC256} & \multicolumn{1}{C{0.90cm}|}{MC512} & \multicolumn{1}{C{0.90cm}|}{AM}                   \\
        \hline
        \hline
        \multicolumn{1}{|C{1.20cm}|}{Airplane} & \multicolumn{1}{C{0.90cm}|}{18.0}       & \multicolumn{1}{C{0.90cm}|}{5.00}  & \multicolumn{1}{C{0.90cm}|}{2.91}           & \multicolumn{1}{C{0.90cm}|}{\textbf{2.86}}  & \multicolumn{1}{C{0.90cm}|}{\textbf{2.86}}  & \multicolumn{1}{C{0.90cm}|}{\textbf{3601}}  & \multicolumn{1}{C{0.90cm}|}{16511} & \multicolumn{1}{C{0.90cm}|}{70051}  & \multicolumn{1}{C{0.90cm}|}{284293} & \multicolumn{1}{C{0.90cm}|}{5948}           & \multicolumn{1}{C{0.90cm}|}{2.54} & \multicolumn{1}{C{0.90cm}|}{3.79}  & \multicolumn{1}{C{0.90cm}|}{14.3}  & \multicolumn{1}{C{0.90cm}|}{159}   & \multicolumn{1}{C{0.90cm}|}{\textbf{2.04}}       \\
        \hline
        \multicolumn{1}{|C{1.20cm}|}{Chair}    & \multicolumn{1}{C{0.90cm}|}{28.4}       & \multicolumn{1}{C{0.90cm}|}{18.7}  & \multicolumn{1}{C{0.90cm}|}{\textbf{15.4}}  & \multicolumn{1}{C{0.90cm}|}{\textbf{15.4}}  & \multicolumn{1}{C{0.90cm}|}{\textbf{15.4}}  & \multicolumn{1}{C{0.90cm}|}{9850}           & \multicolumn{1}{C{0.90cm}|}{41265} & \multicolumn{1}{C{0.90cm}|}{168071} & \multicolumn{1}{C{0.90cm}|}{677782} & \multicolumn{1}{C{0.90cm}|}{\textbf{9465}}  & \multicolumn{1}{C{0.90cm}|}{2.58} & \multicolumn{1}{C{0.90cm}|}{3.86}  & \multicolumn{1}{C{0.90cm}|}{13.1}  & \multicolumn{1}{C{0.90cm}|}{139}   & \multicolumn{1}{C{0.90cm}|}{\textbf{2.57}}       \\
        \hline
        \multicolumn{1}{|C{1.20cm}|}{Rifle}    & \multicolumn{1}{C{0.90cm}|}{12.3}       & \multicolumn{1}{C{0.90cm}|}{7.82}  & \multicolumn{1}{C{0.90cm}|}{6.61}           & \multicolumn{1}{C{0.90cm}|}{\textbf{5.34}}  & \multicolumn{1}{C{0.90cm}|}{5.35}           & \multicolumn{1}{C{0.90cm}|}{\textbf{2109}}  & \multicolumn{1}{C{0.90cm}|}{8885}  & \multicolumn{1}{C{0.90cm}|}{36183}  & \multicolumn{1}{C{0.90cm}|}{146300} & \multicolumn{1}{C{0.90cm}|}{4230}           & \multicolumn{1}{C{0.90cm}|}{2.51} & \multicolumn{1}{C{0.90cm}|}{3.65}  & \multicolumn{1}{C{0.90cm}|}{13.7}  & \multicolumn{1}{C{0.90cm}|}{157}   & \multicolumn{1}{C{0.90cm}|}{\textbf{1.72}}       \\
        \hline
        \multicolumn{1}{|C{1.20cm}|}{Sofa}     & \multicolumn{1}{C{0.90cm}|}{3.10}       & \multicolumn{1}{C{0.90cm}|}{3.06}  & \multicolumn{1}{C{0.90cm}|}{3.05}           & \multicolumn{1}{C{0.90cm}|}{3.05}           & \multicolumn{1}{C{0.90cm}|}{\textbf{3.04}}  & \multicolumn{1}{C{0.90cm}|}{11860}          & \multicolumn{1}{C{0.90cm}|}{48433} & \multicolumn{1}{C{0.90cm}|}{196007} & \multicolumn{1}{C{0.90cm}|}{788885} & \multicolumn{1}{C{0.90cm}|}{\textbf{5214}}  & \multicolumn{1}{C{0.90cm}|}{2.59} & \multicolumn{1}{C{0.90cm}|}{3.99}  & \multicolumn{1}{C{0.90cm}|}{14.9}  & \multicolumn{1}{C{0.90cm}|}{169}   & \multicolumn{1}{C{0.90cm}|}{\textbf{1.85}}       \\
        \hline
        \multicolumn{1}{|C{1.20cm}|}{Table}    & \multicolumn{1}{C{0.90cm}|}{27.1}       & \multicolumn{1}{C{0.90cm}|}{11.4}  & \multicolumn{1}{C{0.90cm}|}{9.35}           & \multicolumn{1}{C{0.90cm}|}{\textbf{9.34}}  & \multicolumn{1}{C{0.90cm}|}{9.35}           & \multicolumn{1}{C{0.90cm}|}{\textbf{9692}}  & \multicolumn{1}{C{0.90cm}|}{42270} & \multicolumn{1}{C{0.90cm}|}{172320} & \multicolumn{1}{C{0.90cm}|}{695810} & \multicolumn{1}{C{0.90cm}|}{10157}          & \multicolumn{1}{C{0.90cm}|}{2.69} & \multicolumn{1}{C{0.90cm}|}{4.28}  & \multicolumn{1}{C{0.90cm}|}{14.2}  & \multicolumn{1}{C{0.90cm}|}{147}   & \multicolumn{1}{C{0.90cm}|}{\textbf{2.46}}       \\
        \hline
      \end{tabular}
    }
  \end{center}
\end{table*}

\subsection{Novel Shape Reconstruction by Learning an Ensemble of Local Decoders}
\label{ExpLocalLearning}

In this section, we investigate whether the results of AM can be improved by learning an ensemble of local decoders for novel shape reconstruction. We train a hypernetwork of MLP with depth 2 and width 1,024, which learns weights for an ensemble of 4 subnetworks for occupancy modeling; each subnetwork is of depth 4 and width 32; thus in total the ensemble has 512 neurons, slightly more than the D6-W60 model used in global decoding. We again conduct experiments on the five ShapeNet categories by training the hypernetwork on the training shapes and testing on novel ones. Other learning setups are given in the beginning of Section \ref{SecExp}. Quantitative results in Table \ref{TableExpsLocalShapeLearning} show that the ensemble model indeed improves the accuracies of mesh reconstructions over those obtained by a single, global model, at the cost of a slightly slower inference. On all the five categories, AM, \emph{after mesh simplification at a ratio of 10\%}, gives more accurate reconstructions at a much faster inference and much fewer numbers of mesh faces than MC512 does. The qualitative comparisons given in Appendix \ref{AppendixNovelShapeReconComp} are consistent with the quantitative ones, where the recovered components by individual decoders of the ensemble suggest that the ensemble learns the shapes naturally in a constructive manner.

\begin{table*}[!t]
  \caption{Quantitative comparisons between analytic marching (AM) and marching cubes (MC) in the context of learning an ensemble of local decoders for reconstructions of novel shape instances. Experiments are conducted on shape instances of five categories from ShapeNet, by training a hypernetwork that gives weights of the ensemble for occupancy modeling; the ensemble is formed by aggregating, via max pooling, outputs of 4 subnetworks, each of which is of depth 4 and width 32. \emph{Results of AM are after simplification of the originally recovered meshes at a ratio of 10\%}. For MC, the sampling resolutions of 3D points range from $64^3$ to $512^3$.  } \vspace{-0.2cm}
  \label{TableExpsLocalShapeLearning}
  \vspace{-0.3cm}
  \begin{center}
    \scalebox{0.8}{
      \begin{tabular}{cccccccccccccccc}
        \hline
        \multicolumn{1}{|C{1.20cm}|}{Metric}   & \multicolumn{5}{c|}{CD}                                                                                                                                                                                            & \multicolumn{5}{c|}{\textsharp TriFace}                                                                                                                                                           & \multicolumn{5}{c|}{Time (sec.)}                                                                                                                                                                 \\
        \cline{1-16}
        \multicolumn{1}{|C{1.20cm}|}{Method}   &  \multicolumn{1}{C{0.90cm}|}{MC64}       & \multicolumn{1}{C{0.90cm}|}{MC128} & \multicolumn{1}{C{0.90cm}|}{MC256}    & \multicolumn{1}{C{0.90cm}|}{MC512}          & \multicolumn{1}{C{0.90cm}|}{AM}              & \multicolumn{1}{C{0.90cm}|}{MC64}  & \multicolumn{1}{C{0.90cm}|}{MC128} & \multicolumn{1}{C{0.90cm}|}{MC256}  & \multicolumn{1}{C{0.90cm}|}{MC512}  & \multicolumn{1}{C{0.90cm}|}{AM}             & \multicolumn{1}{C{0.90cm}|}{MC64} & \multicolumn{1}{C{0.90cm}|}{MC128} & \multicolumn{1}{C{0.90cm}|}{MC256} & \multicolumn{1}{C{0.90cm}|}{MC512} & \multicolumn{1}{C{0.90cm}|}{AM}               \\
        \hline
        \hline
        \multicolumn{1}{|C{1.20cm}|}{Airplane} &  \multicolumn{1}{C{0.90cm}|}{2.17}       & \multicolumn{1}{C{0.90cm}|}{0.803} & \multicolumn{1}{C{0.90cm}|}{0.608}    & \multicolumn{1}{C{0.90cm}|}{\textbf{0.607}} & \multicolumn{1}{C{0.90cm}|}{\textbf{0.607}}  & \multicolumn{1}{C{0.90cm}|}{\textbf{4881}}  & \multicolumn{1}{C{0.90cm}|}{19626} & \multicolumn{1}{C{0.90cm}|}{78375}  & \multicolumn{1}{C{0.90cm}|}{311699} & \multicolumn{1}{C{0.90cm}|}{6775}  & \multicolumn{1}{C{0.90cm}|}{\textbf{3.97}} & \multicolumn{1}{C{0.90cm}|}{6.75}  & \multicolumn{1}{C{0.90cm}|}{28.6}  & \multicolumn{1}{C{0.90cm}|}{198}   & \multicolumn{1}{C{0.90cm}|}{4.62}    \\
        \hline
        \multicolumn{1}{|C{1.20cm}|}{Chair}    &  \multicolumn{1}{C{0.90cm}|}{3.95}       & \multicolumn{1}{C{0.90cm}|}{3.82}  & \multicolumn{1}{C{0.90cm}|}{3.55}     & \multicolumn{1}{C{0.90cm}|}{\textbf{3.51}}  & \multicolumn{1}{C{0.90cm}|}{\textbf{3.51}}   & \multicolumn{1}{C{0.90cm}|}{12538} & \multicolumn{1}{C{0.90cm}|}{47807} & \multicolumn{1}{C{0.90cm}|}{184007} & \multicolumn{1}{C{0.90cm}|}{725064} & \multicolumn{1}{C{0.90cm}|}{\textbf{5625}}  & \multicolumn{1}{C{0.90cm}|}{3.97} & \multicolumn{1}{C{0.90cm}|}{6.72}  & \multicolumn{1}{C{0.90cm}|}{27.2}  & \multicolumn{1}{C{0.90cm}|}{186}   & \multicolumn{1}{C{0.90cm}|}{\textbf{3.61}}    \\
        \hline
        \multicolumn{1}{|C{1.20cm}|}{Rifle}    &  \multicolumn{1}{C{0.90cm}|}{2.48}       & \multicolumn{1}{C{0.90cm}|}{1.55}  & \multicolumn{1}{C{0.90cm}|}{1.48}     & \multicolumn{1}{C{0.90cm}|}{\textbf{1.42}}  & \multicolumn{1}{C{0.90cm}|}{\textbf{1.42}}   & \multicolumn{1}{C{0.90cm}|}{\textbf{2683}}  & \multicolumn{1}{C{0.90cm}|}{10132} & \multicolumn{1}{C{0.90cm}|}{38553}  & \multicolumn{1}{C{0.90cm}|}{150220} & \multicolumn{1}{C{0.90cm}|}{3609}  & \multicolumn{1}{C{0.90cm}|}{3.91} & \multicolumn{1}{C{0.90cm}|}{6.57}  & \multicolumn{1}{C{0.90cm}|}{26.8}  & \multicolumn{1}{C{0.90cm}|}{186}   & \multicolumn{1}{C{0.90cm}|}{\textbf{3.16}}    \\
        \hline
        \multicolumn{1}{|C{1.20cm}|}{Sofa}     &  \multicolumn{1}{C{0.90cm}|}{2.03}       & \multicolumn{1}{C{0.90cm}|}{1.97}  & \multicolumn{1}{C{0.90cm}|}{1.91}     & \multicolumn{1}{C{0.90cm}|}{1.87}           & \multicolumn{1}{C{0.90cm}|}{\textbf{1.86}}   & \multicolumn{1}{C{0.90cm}|}{14237} & \multicolumn{1}{C{0.90cm}|}{52950} & \multicolumn{1}{C{0.90cm}|}{203497} & \multicolumn{1}{C{0.90cm}|}{800701} & \multicolumn{1}{C{0.90cm}|}{\textbf{4100}}  & \multicolumn{1}{C{0.90cm}|}{3.99} & \multicolumn{1}{C{0.90cm}|}{6.82}  & \multicolumn{1}{C{0.90cm}|}{28.5}  & \multicolumn{1}{C{0.90cm}|}{195}   & \multicolumn{1}{C{0.90cm}|}{\textbf{3.49}}    \\
        \hline
        \multicolumn{1}{|C{1.20cm}|}{Table}    &  \multicolumn{1}{C{0.90cm}|}{6.36}       & \multicolumn{1}{C{0.90cm}|}{3.89}  & \multicolumn{1}{C{0.90cm}|}{3.66}     & \multicolumn{1}{C{0.90cm}|}{3.63}           & \multicolumn{1}{C{0.90cm}|}{\textbf{3.62}}   & \multicolumn{1}{C{0.90cm}|}{10389} & \multicolumn{1}{C{0.90cm}|}{43442} & \multicolumn{1}{C{0.90cm}|}{174667} & \multicolumn{1}{C{0.90cm}|}{700780} & \multicolumn{1}{C{0.90cm}|}{\textbf{4951}}  & \multicolumn{1}{C{0.90cm}|}{4.10} & \multicolumn{1}{C{0.90cm}|}{7.62}  & \multicolumn{1}{C{0.90cm}|}{34.5}  & \multicolumn{1}{C{0.90cm}|}{247}   & \multicolumn{1}{C{0.90cm}|}{\textbf{3.71}}    \\
        \hline
      \end{tabular}
    }
  \end{center}
\end{table*}


\section{Conclusion and Future Research}

We have presented in this paper an exact meshing solution from deep implicit surface networks. Our analytic marching algorithm works by marching along the analytic cells in the input 3D space and exactly recovering the analytic faces. We have proved that under mild, numerical conditions, the recovered analytic faces are guaranteed to connect and form a closed, piecewise planar surface. Our theory and algorithm also support advanced MLPs with shortcut connections and max pooling. We have empirically demonstrated the advantages of our method over existing meshing algorithms in the contexts of either direct shape encoding or learning to decode novel shape instances.

Our current theory and algorithm assume the MLPs with ReLU activations. For other variants of the ReLU family, such as leaky ReLU \cite{Maas2013} or parametric ReLU \cite{He2015}, we note that it is possible to develop the corresponding theory based on the recent result in \cite{Hu2020PWLNet}, which extends \cite{Montufar2014} and bounds the numbers of linear regions partitioned by general, piece-wise linear networks. For softer activations such as Softplus \cite{Lange2014}, ELU \cite{Clevert2016}, or Swish \cite{Ramachandran2017}, one may employ them in an asymptotic manner during training of the implicit networks, similar to the efficient training scheme used in Section \ref{SecDirectShapeEncoding}; after training, the networks approach ReLU based ones, whose zero-level isosurfaces can then be recovered by analytic marching. For more general smooth activations, including the very recent sinusoid one \cite{Sitzmann2019}, it remains unclear to develop their exact meshing theories. One may consider using piecewise linear functions to approximate the smooth activations, or employing a strategy of distillation \cite{KnowledgeDistill} to transfer the knowledge learned in the teacher implicit networks of general smooth activations into the student, ReLU based ones. We leave these extended studies as future research.

Direct shape encoding presented in Section \ref{SecDirectShapeEncoding} suggests a possibly more compact way of coding, storing, and transmitting 3D shapes as implicit surface networks, as pointed out firstly in \cite{davies2021effectiveness}. However, the neurally encoded shapes in \cite{davies2021effectiveness} are still to be decoded via meshing algorithms such as marching cubes, which, as we have argued, would cause loss of the precisions already encoded in the implicit networks. As an algorithm of exact meshing, our proposed analytic marching makes a solid step towards making implicit surface networks truly as a new promise of compact shape representations.



%

%
%

\ifCLASSOPTIONcaptionsoff
  \newpage
\fi



%




\bibliographystyle{IEEEtran}
\bibliography{references}


\begin{IEEEbiography}[{\includegraphics[width=1in,height=1.25in,clip,keepaspectratio]{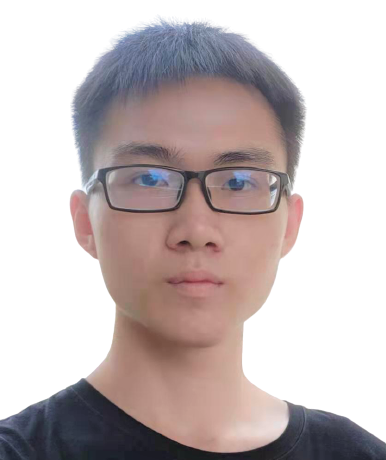}}]{Jiabao Lei}
 is currently working toward the master degree in the School of Electronic and Information Engineering, South China University of Technology, Guangzhou, China. Recently his research interests mainly include 3D geometric representation and surface reconstruction in the field of deep learning.
\end{IEEEbiography}

\begin{IEEEbiography}[{\includegraphics[width=1in,height=1.25in,clip,keepaspectratio]{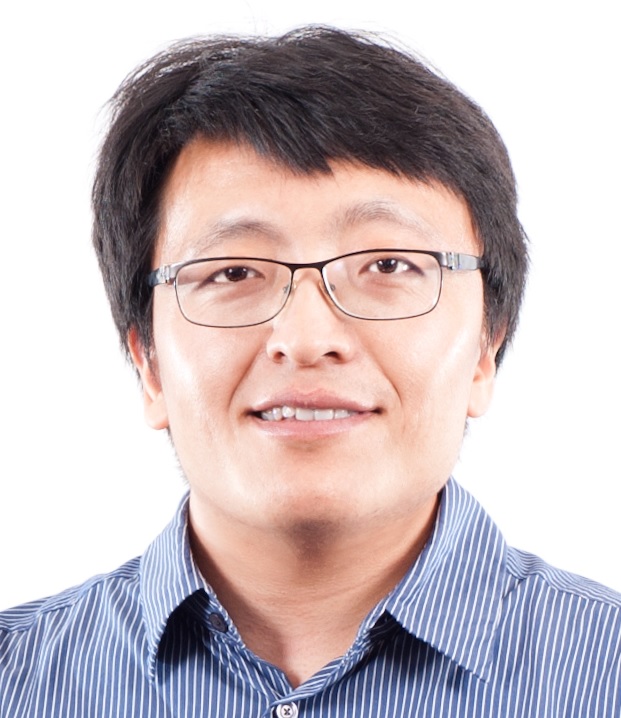}}]{Kui Jia}
	received the Ph.D. degree in computer science from the Queen Mary University of London, London, U.K., in 2007. He was with the Shenzhen Institute of Advanced Technology of the Chinese Academy of Sciences, Shenzhen, China, Chinese University of Hong Kong, Hong Kong, the Institute of Advanced Studies, University of Illinois at Urbana-Champaign, Champaign, IL, USA, and the University of Macau, Macau, China. He is currently a Professor with the School of Electronic and Information Engineering, South China University of Technology, Guangzhou, China, and is the Director of Geometric Perception and Intelligence Research Lab. His recent research focuses on theoretical deep learning and its applications in vision and robotic problems, including deep learning of 3D data and deep transfer learning. He has been serving as Associate Editors for TIP and TSMC.
\end{IEEEbiography}

\begin{IEEEbiography}[{\includegraphics[width=1in,height=1.25in,clip,keepaspectratio]{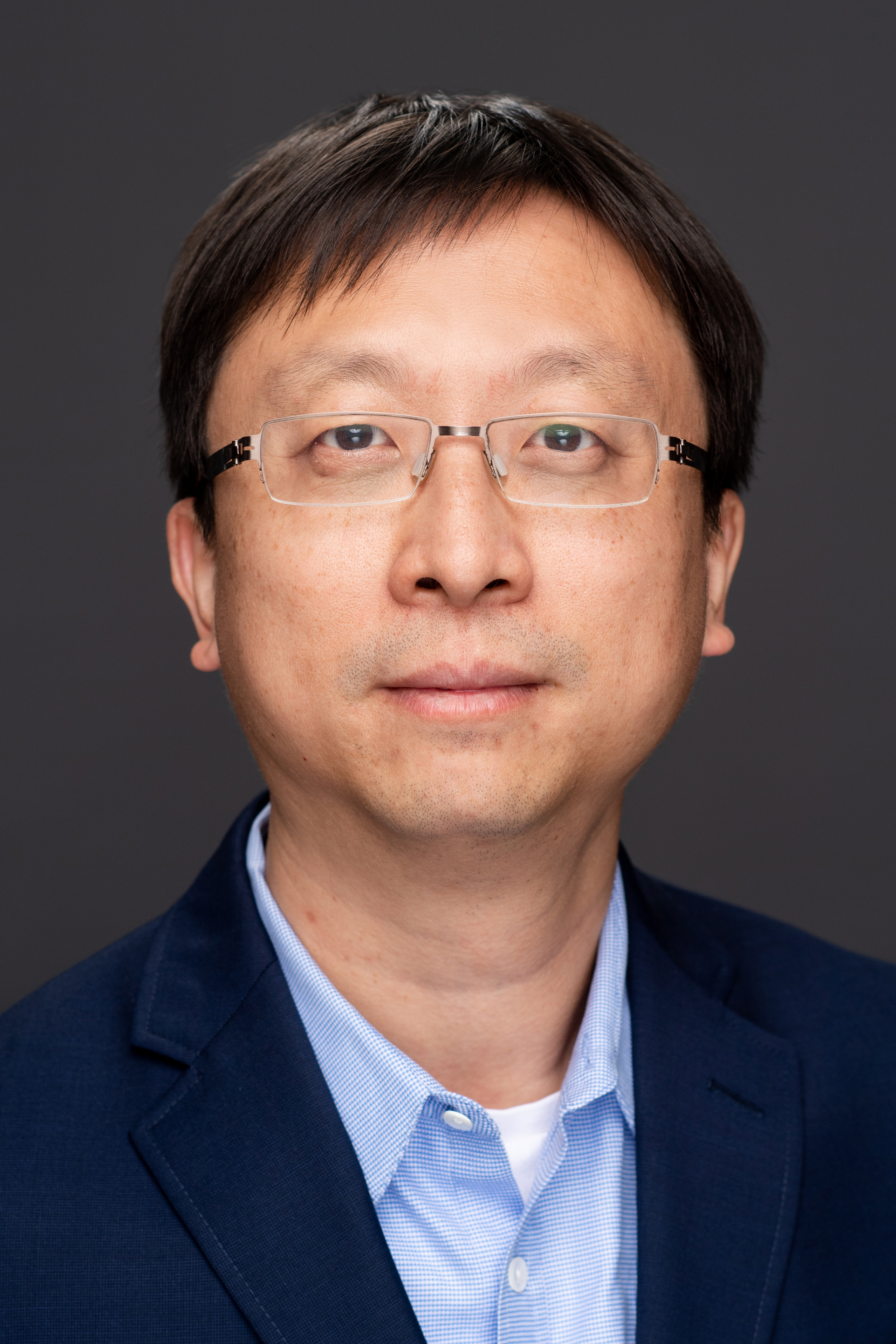}}]{Yi Ma}
  received the bachelor’s degree in automation
  and applied mathematics from Tsinghua University,
  Beijing, China, in 1995, and the master’s degree
  in EECS and mathematics and the Ph.D. degree in
  EECS from the University of California at Berkeley,
  Berkeley, in 2000. From 2000 to 2011, he has served
  on the Faculty of the ECE Department, University
  of Illinois at Urbana–Champaign. From 2009 to
  2014, he has served as a Principal Researcher
  and a Research Manager for the Visual Computing
  Group, Microsoft Research Asia, Beijing, China.
  From 2014 to 2017, he was a Professor and the Executive Dean of the School
  of Information Science and Technology, ShanghaiTech University, China.
  Since January 2018, he has been with the Faculty of the EECS Department,
  University of California at Berkeley. He has written two textbooks An
  Invitation to 3D Vision (Springer) and Generalized Principal Component
  Analysis (Springer). He is a fellow of ACM. He has received best paper
  awards from ICCV, ECCV, and ACCV. He received the CAREER Award
  from the NSF and the YIP Award from the ONR. He has served as an
  Associate Editor for IJCV, SIIMS, SIMODS, the IEEE TRANSACTIONS ON
  PATTERN ANALYSIS AND MACHINE INTELLIGENCE (PAMI), and the IEEE
  TRANSACTIONS ON INFORMATION THEORY.
\end{IEEEbiography}



\newpage

\appendices

\section{Proof of Theorem \ref{MainResult}}
\label{AppendixMainResultProof}

\begin{proof}
  Since $F = f\circ\mathbf{T}$ is constructed from a ReLU based MLP $\bm{T}$, its zero-level isosurface $\mathcal{Z}$ is piecewise planar. The proof proceeds by first showing that each planar face on $\mathcal{Z}$ uniquely corresponds to an analytic face in an analytic cell, and then showing that for any pair of planar faces connected on $\mathcal{Z}$ (since $\mathcal{Z}$ is closed by assumption), their corresponding analytic faces are connected via boundaries of their respective analytic cells.

  Let $\mathcal{P}_1$ denote a planar face on the surface $\mathcal{Z}$ 
  , and $\mathbf{n}_1 \in \mathbb{R}^3$ be its normal. We have $\mathbf{n}_1^{\top}\mathbf{x} = 0 \ \forall \ \mathbf{x}\in \mathcal{P}_1$. Equation (\ref{EqnSDFPlaneFunctional}) tells that $\mathbf{n}_1$ must be proportional at least to one of $\{\mathbf{a}_F^r | r\in\mathcal{R} \}$. By the unique plane condition, i.e., each of $\{\mathbf{a}_F^r | r\in\mathcal{R} \}$ is uniquely defined, we have $\mathbf{n}_1 \propto \mathbf{a}_F^{r_1\top}$ of a certain region $r_1$. Assume $r_1$ is not an analytic cell, which suggests that there exists no intersection between $\mathbf{a}_F^{r_1}$ and $r_1$ and we have $\mathbf{a}_F^{r_1}\mathbf{x} = \mathbf{n}_1^{\top}\mathbf{x} \neq 0$ for all $\mathbf{x} \in r_1$, and thus $\mathcal{P}_1 \land r_1 = \emptyset$; it suggests that the normal $\mathbf{n}_1$ of $\mathcal{P}_1$ is induced in a different region $r_1'$ by $\mathbf{n}_1 \propto \mathbf{a}_F^{r_1'} = \mathbf{w}_f^{\top}\mathbf{T}^{r_1'}$, which contradicts with the assumed unique plane condition. We thus have that $r_1$ must be an analytic cell.

  Let $\mathbf{n}_1 \propto \mathbf{a}_F^{\tilde{r}_1\top}$ of a certain analytic cell $\tilde{r}_1 \in \widetilde{\mathcal{R}}$ (or $\mathcal{C}_F^{\tilde{r}_1}$), and we have the analytic face $\mathcal{P}_F^{\tilde{r}_1} \subseteq \mathcal{P}_1$. Assume there exist $\mathcal{P}_1 - \mathcal{P}_F^{\tilde{r}_1} = \{ \mathbf{x} \in \mathcal{Z} | \mathbf{x} \in \mathcal{P}_1, \mathbf{x} \notin \mathcal{P}_F^{\tilde{r}_1} \}$, which means that for any $\mathbf{x} \in \mathcal{P}_1 - \mathcal{P}_F^{\tilde{r}_1}$, it resides in an analytic face $\mathcal{P}_F^{\tilde{r}_1'}$ of a different cell $\mathcal{C}_F^{\tilde{r}_1'}$; since $\mathbf{x} \in \mathcal{P}_1$, we have $\mathbf{n}_1 \propto \mathbf{a}_F^{\tilde{r}_1'\top}$ and thus $\mathbf{a}_F^{\tilde{r}_1} \propto \mathbf{a}_F^{\tilde{r}_1'}$, which contradicts with the unique plane condition of $\mathbf{a}_F^{\tilde{r}_1} \not\propto \mathbf{a}_F^{\tilde{r}_1'}$.
  We thus have $\mathcal{P}_1 = \mathcal{P}_F^{\tilde{r}_1}$ and $\mathcal{P}_1 \subset \mathcal{C}_F^{\tilde{r}_1}$. By the definition (\ref{EqnAnalyticFace}) of analytic face, the above argument also tells that planar faces on $\mathcal{Z}$ and analytic faces $\{ \mathcal{C}_F^{\tilde{r}} | \tilde{r} \in \widetilde{\mathcal{R}} \}$ are one-to-one corresponded.

  Assume $\mathcal{P}_1$ connects with another planar face $\mathcal{P}_2$ on a shared edge segment $\mathcal{E} = \{ \mathbf{x} \in \mathcal{Z} | \mathbf{x} \in \mathcal{P}_1, \mathbf{x} \in \mathcal{P}_2\}$. Define the normal of $\mathcal{P}_2$ as $\mathbf{n}_2 \in \mathbb{R}^3$, we have $\mathbf{n}_1 \not\propto \mathbf{n}_2$. Let $\mathcal{P}_2 \subset \mathcal{C}_F^{\tilde{r}_2}$, and we thus have $\mathcal{E} \subset \mathcal{C}_F^{\tilde{r}_1}$ and $\mathcal{E} \subset \mathcal{C}_F^{\tilde{r}_2}$, which tells that the two cells $\mathcal{C}_F^{\tilde{r}_1}$ and $\mathcal{C}_F^{\tilde{r}_2}$ connect at least on $\mathcal{E}$. Due to the monotonous and convex nature of linear analytic cells $\{ \mathcal{C}_F^{\tilde{r}} | \tilde{r} \in \widetilde{\mathcal{R}} \}$,  $\mathcal{E}$ must be on the boundaries of both $\mathcal{C}_F^{\tilde{r}_1}$ and $\mathcal{C}_F^{\tilde{r}_2}$, and the boundaries of $\mathcal{C}_F^{\tilde{r}_1}$ and $\mathcal{C}_F^{\tilde{r}_2}$ share at least on $\mathcal{E}$. There exist two cases for the connection of cell boundaries on $\mathcal{E}$: 1) in the general case, $\mathcal{C}_F^{\tilde{r}_1}$ and $\mathcal{C}_F^{\tilde{r}_2}$ share a boundary $\mathcal{B}_F^{\tilde{r}_1\tilde{r}_2}$ defined by a hyperplane $\bm{H}_{lk}^{\tilde{r}_1\tilde{r}_2} = \{ \mathbf{x} \in \mathbb{R}^3 | \mathbf{a}^{\tilde{r}_1\tilde{r}_2}_{lk}\mathbf{x} = 0 \}$, and we have $\mathcal{E} \in \mathcal{B}_F^{\tilde{r}_1\tilde{r}_2}$, which, based on Corollary \ref{LemmaRegionAssociatedNeuronMapping} and Definition \ref{DefinitionNeuronMLPState}, suggests that the two cells have a switching neuron state $s_{lk}(\mathbf{x}) \ \forall \ \mathbf{x} \in \mathcal{B}_F^{\tilde{r}_1\tilde{r}_2}$, and consequently a switching neuron state $s_{lk}(\mathbf{x}) \ \forall \ \mathbf{x} \in \mathcal{E}$; 2) in some rare case, $\mathcal{E}$ coincides with a cell edge of $\mathcal{C}_F^{\tilde{r}_1}$ defined by $\{ \mathbf{x} \in \mathbb{R}^3 | \mathbf{a}^{\tilde{r}_1}_{l_1 k_1}\mathbf{x} = 0, \mathbf{a}^{\tilde{r}_1}_{l_1' k_1'}\mathbf{x} = 0 \}$,  and a cell edge of $\mathcal{C}_F^{\tilde{r}_2}$ defined by $\{ \mathbf{x} \in \mathbb{R}^3 | \mathbf{a}^{\tilde{r}_2}_{l_2 k_2}\mathbf{x} = 0, \mathbf{a}^{\tilde{r}_2}_{l_2' k_2'}\mathbf{x} = 0 \}$, and it is not necessary that $l_1k_1$ and $l_2k_2$ specify the same neuron, and $l_1'k_1'$ and $l_2'k_2'$ specify another same neuron. Due to a phenomenon similar to the blessing of (high) dimensionality \cite{Gorban2018}, the second case of coincidence is expected to happen with a low probability. In any of the two cases, the boundaries $\mathcal{C}_F^{\tilde{r}_1}$ and $\mathcal{C}_F^{\tilde{r}_2}$ respectively associated with $\mathcal{P}_1$ and $\mathcal{P}_2$ connect on $\mathcal{E}$.

  Since for any pair of planar faces $\mathcal{P}_1$ and $\mathcal{P}_2$ connected on $\mathcal{Z}$, we prove that they are uniquely corresponded to a pair of analytic faces $\mathcal{P}_F^{\tilde{r}_1}$ and $\mathcal{P}_F^{\tilde{r}_2}$, which are polygon faces connected via boundaries of their respective analytic cells $\mathcal{C}_F^{\tilde{r}_1}$ and $\mathcal{C}_F^{\tilde{r}_2}$. The theorem is proved.
\end{proof}

\section{Algorithmic Details of Pivoting Enumeration}
\label{AppendixPivotingEnum}

In this section, we present the Algorithm \ref{AlgPivotingEnumeration} of pivoting enumeration that greatly improves the efficiency of analytic marching.

\begin{algorithm}
  \caption{Pivoting enumeration in analytic marching.}
  \label{AlgPivotingEnumeration}
  \begin{flushleft}
    \textbf{INPUT:} A point $\boldsymbol{x} \in \mathcal{P}_F^{\tilde{r}}$; a hyperplane $\bm{H}_{l k}^{\tilde{r}} \in \{ \bm{H}_{lk}^{\tilde{r}}\}$ that is a true boundary of the working cell $\mathcal{C}_F^{\tilde{r}}$. \\
    \textbf{OUTPUT:} A set $\mathcal{V}_{\mathcal{P}}^{\tilde{r}}$ of ordered vertices.
  \end{flushleft}

  \begin{algorithmic}[1]

    \State \algmultiline{%
        Establish an index set $\mathcal{I}^{\tilde{r}} = \{ (l_1, k_1), \dots, (l_{N}, k_{N}) \}$ by sorting, in an increasing order, the Euclidean distances from each of the $N$ hyperplanes in $\{ \bm{H}_{lk}^{\tilde{r}}\}$ to the point $\boldsymbol{x} \in \mathcal{P}_F^{\tilde{r}}$, where $N = n_1 + \dots + n_L$ is the total number of neurons in $\bm{T}$.
    }
    \State \algmultiline{%
        Introduce the auxiliary $t, t', t'' \in \mathbb{N}$; use $(l_t, k_t)$ to index the firstly identified boundary plane $\bm{H}_{l k}^{\tilde{r}}$, written as $\bm{H}_{l_t k_t}^{\tilde{r}}$ (this same boundary is denoted as $(l_{*}, k_{*})$ and $\bm{H}_{l_{*} k_{*}}^{\tilde{r}}$ for a later reference). Let $\mathcal{I}_{/_t}^{\tilde{r}} = \{ (l_1, k_1), \dots, (l_{t-1}, k_{t-1}), (l_{t+1}, k_{t+1}), \dots, (l_{N}, k_{N}) \}$, and $\mathcal{I}_{/_{t'}/_{t}}^{\tilde{r}} = \{ (l_1, k_1), \dots, (l_{t'-1}, k_{t'-1}), (l_{t'+1}, k_{t'+1}), $ $\dots, (l_{t-1}, k_{t-1}), (l_{t+1}, k_{t+1}), \dots, (l_{N}, k_{N}) \}$.
    }
    \State \algmultiline{%
      Initialize $\mathcal{V}_{\mathcal{P}}^{\tilde{r}} = \emptyset$ and $t'' = 1$.
    }
    \While{$\mathcal{V}_{\mathcal{P}}^{\tilde{r}} = \emptyset$}
          \State \algmultiline{%
            Solve the system of equations $[\mathbf{a}_{l_{t} k_{t}}^{\tilde{r}}; \mathbf{a}_{l_{t''} k_{t''}}^{\tilde{r}}; \mathbf{a}_F^{\tilde{r}}] \mathbf{x} = \bm{0}$ to have a vertex candidate $\bm{v} \in \mathbb{R}^3$.
          }
          \If{$\bm{v}$ satisfies condition (\ref{EqnAnalyticCellSystem})}
              \State \algmultiline{
                  Push $\bm{v}$ into $\mathcal{V}_{\mathcal{P}}^{\tilde{r}}$.
              }
              \State \algmultiline{
                  Update $(l_{t'}, k_{t'}) = (l_{t}, k_{t})$.
              }
              \State \algmultiline{
                  Update $(l_t, k_t) = (l_{t''}, k_{t''})$.
              }
          \Else
              \State \algmultiline{
                  Update $t'' \leftarrow t'' + 1$ upon $(l_{t''+1}, k_{t''+1}) \in \mathcal{I}_{/_t}^{\tilde{r}}$ or $t'' \leftarrow t'' + 2$ upon $(l_{t''+2}, k_{t''+2}) \in \mathcal{I}_{/_t}^{\tilde{r}}$.
              }
          \EndIf
    \EndWhile
    \While{\textbf{true}}
        \State \algmultiline{
            Solve the system of equations $[\mathbf{a}_{l_{t} k_{t}}^{\tilde{r}}; \mathbf{a}_{l_{t''} k_{t''}}^{\tilde{r}}; \mathbf{a}_F^{\tilde{r}}] \mathbf{x} = \bm{0}$ to have a vertex candidate $\bm{v} \in \mathbb{R}^3$.
        }
        \If{$\bm{v}$ satisfies condition (\ref{EqnAnalyticCellSystem})}
              \State \algmultiline{
                  Push $\bm{v}$ into $\mathcal{V}_{\mathcal{P}}^{\tilde{r}}$.
              }
              \If{$(l_{t''}, k_{t''})$ indexes $\bm{H}_{l_{*} k_{*}}^{\tilde{r}}$}
                \State \textbf{break}
              \Else
                \State \algmultiline{
                  Update $(l_{t'}, k_{t'}) = (l_{t}, k_{t})$.
                }
                \State \algmultiline{
                  Update $(l_t, k_t) = (l_{t''}, k_{t''})$.
                }
                \State \algmultiline{
                  Update $\mathcal{I}_{/_{t'}/_t}^{\tilde{r}} = \{ (l_1, k_1), \dots, (l_{t'-1}, k_{t'-1}), $ $ (l_{t'+1}, k_{t'+1}), \dots, (l_{t-1}, k_{t-1}), (l_{t+1}, k_{t+1}), \dots, $ $ (l_{N}, k_{N}) \}$.
                }
                \State \algmultiline{
                  Reset $t'' = 1$.
                }
              \EndIf
          \Else
              \State \algmultiline{
                  Update $t'' \leftarrow t'' + 1$ upon $(l_{t''+1}, k_{t''+1}) \in \mathcal{I}_{/_{t'}/_t}^{\tilde{r}}$ or $t'' \leftarrow t'' + 2$ upon $(l_{t''+2}, k_{t''+2}) \in \mathcal{I}_{/_{t'}/_t}^{\tilde{r}}$.
              }
          \EndIf
    \EndWhile

  \end{algorithmic}
\end{algorithm}

\section{Additional Results of the Richly Detailed Dataset}
\label{AppendixRDAdditionalResults}

\begin{figure*}[!htbp]
  \vskip 0.1in
  \begin{center}
    \hspace{0.0cm}\includegraphics[scale=0.084]{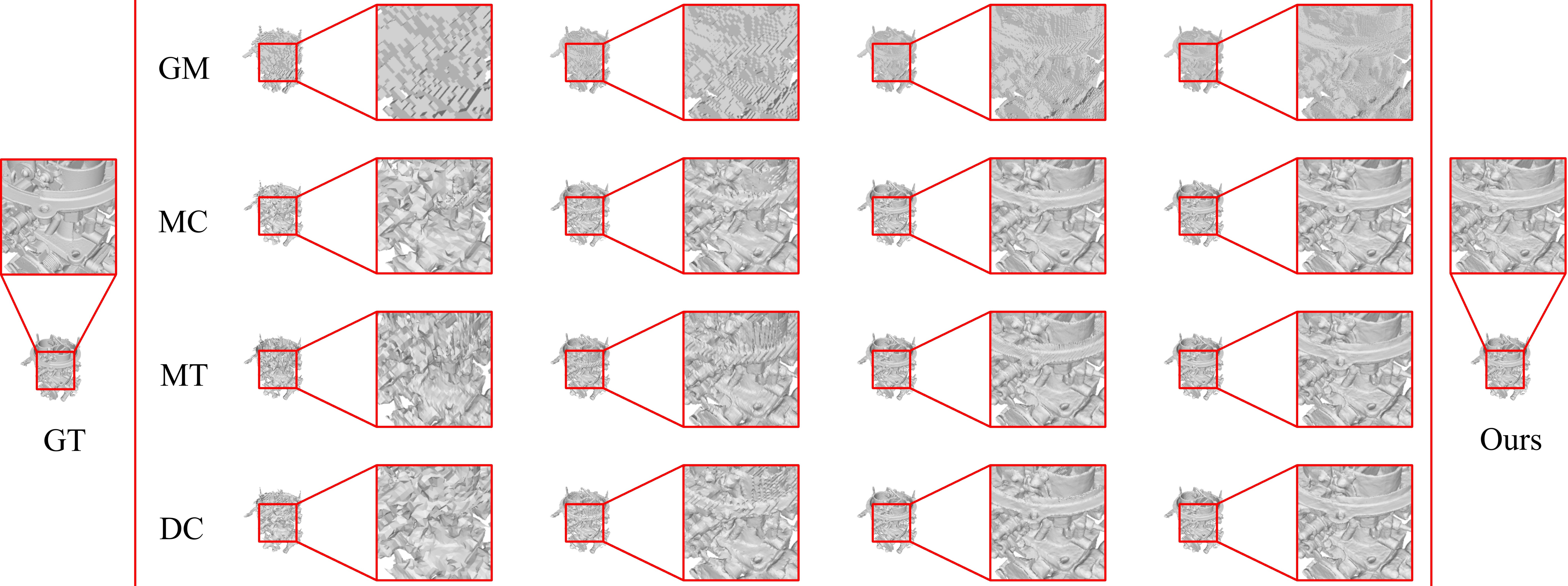}
    \vspace{0.2cm}

    \hspace{0.0cm}\includegraphics[scale=0.084]{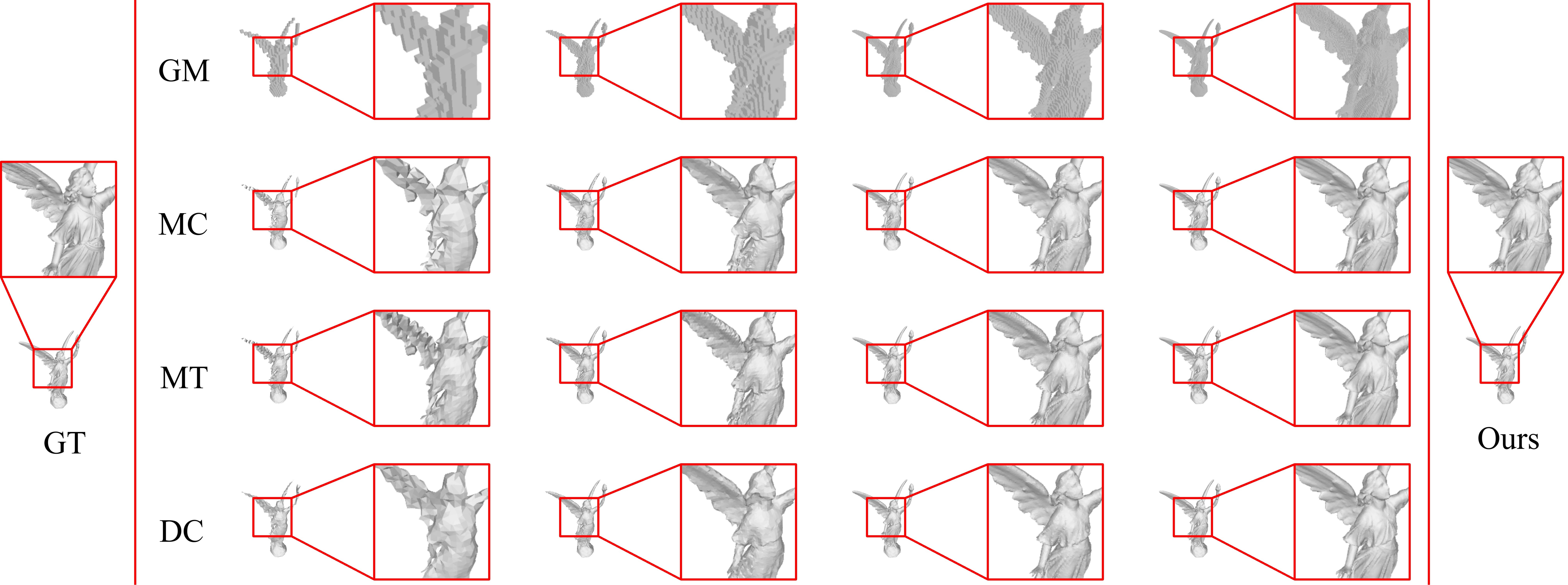}
    \vspace{0.2cm}

    \hspace{0.0cm}\includegraphics[scale=0.084]{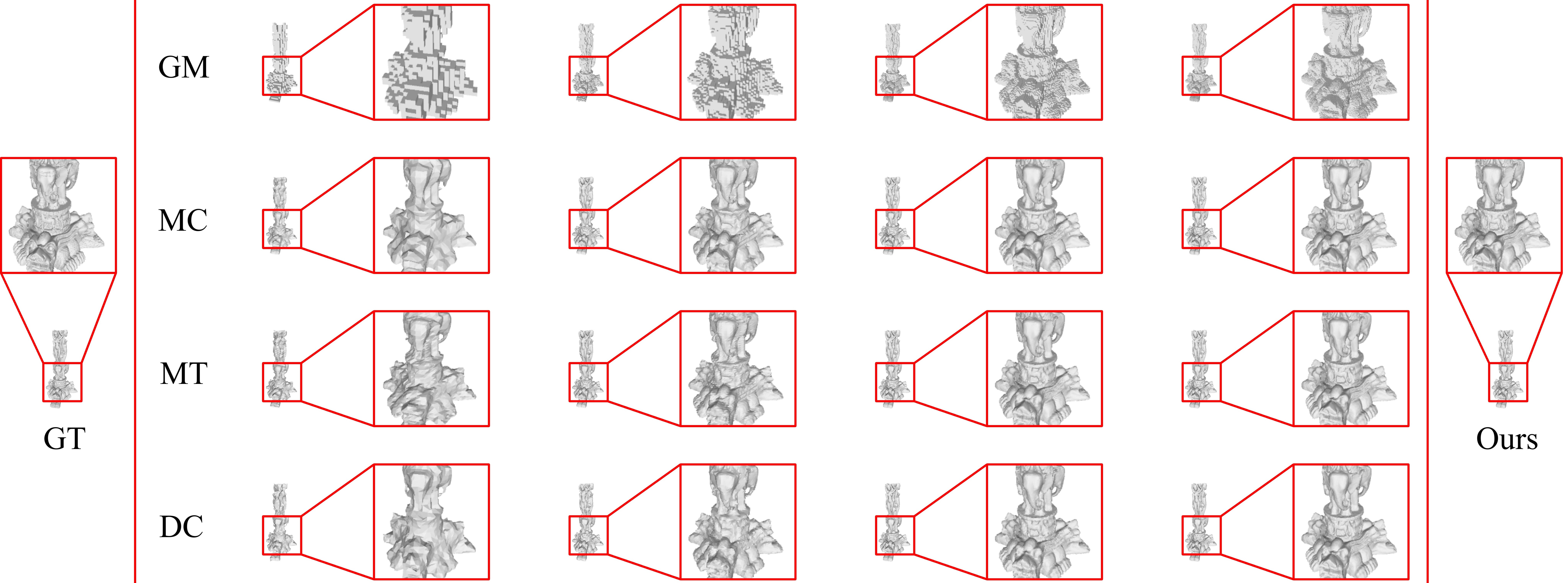}
    \vspace{0.00cm}

    \hspace{1.8cm}\scalebox{1.05}
    {
      \begin{tabular}{C{3.0cm} C{3.0cm} C{3.0cm} C{3.0cm}}
        MC64 &  MC128 &  MC256 & MC512
      \end{tabular}
    }

    \caption{Qualitative comparisons of direct shape encoding. The shown shape instances are fitted to an MLP of depth 8 and width 60 for SDF modeling. For greedy meshing (GM), marching cubes (MC), marching tetrahedra (MT), and dual contouring (DC), results under a resolution range of discrete point sampling from $64^3$ to a GPU memory limit of $512^3$ are presented.  }
    \label{FigExpsDirectShapeEncodingMoreVisualizations}
  \end{center}

  \vskip -0.1in
\end{figure*}

We show additional results of direct shape encoding for the Richly Detailed (RD) dataset in Fig. \ref{FigExpsDirectShapeEncodingMoreVisualizations}. 

\section{Qualitative Results of Novel Shape Reconstructions}
\label{AppendixNovelShapeReconComp}

\begin{figure*}[!htbp]
  \vskip 0.1in
  \begin{center}
    \includegraphics[scale=0.12]{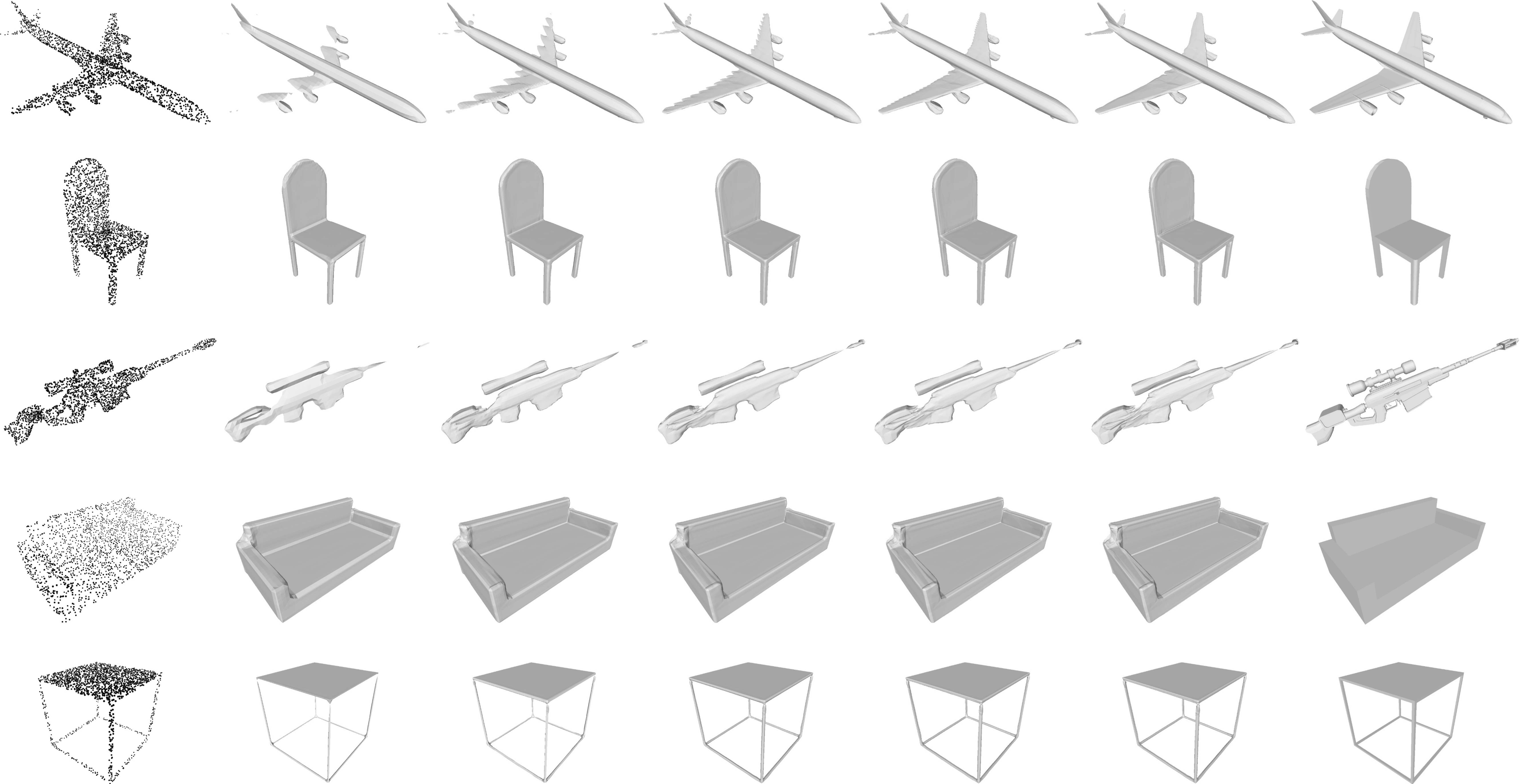}

      \vspace{0.2cm}\hspace{0.1cm}
      \begin{tabular}{p{1.9cm}p{1.9cm}p{1.9cm}p{1.9cm}p{1.9cm}p{1.9cm}p{1.9cm}}
        Input PC &
        \hspace{0.2cm}MC64 &
        \hspace{0.3cm}MC128 &
        \hspace{0.35cm}MC256 &
        \hspace{0.45cm}MC512 &
        \hspace{0.6cm}Ours &
        \hspace{0.8cm}GT \\
      \end{tabular}

    \caption{Qualitative comparisons between analytic marching (AM) and marching cubes (MC) in the context of learning a global model for decoding of novel shape instances. Experiments are conducted on shape instances of five categories from ShapeNet, by training a hypernetwork that gives weights of a light MLP with depth 6 and width 60 for SDF modeling. We show an example from each of the five categories. \emph{Results of AM are after simplification of the originally recovered meshes at a ratio of 10\%}. For MC, the sampling resolutions of 3D points range from $64^3$ to $512^3$.  }
    \label{FigExpsGlobalShapeLearning}
  \end{center}
  \vskip -0.1in
\end{figure*}

\begin{figure*}[!htbp]
  \vskip 0.1in
  \begin{center}
    \includegraphics[scale=0.12]{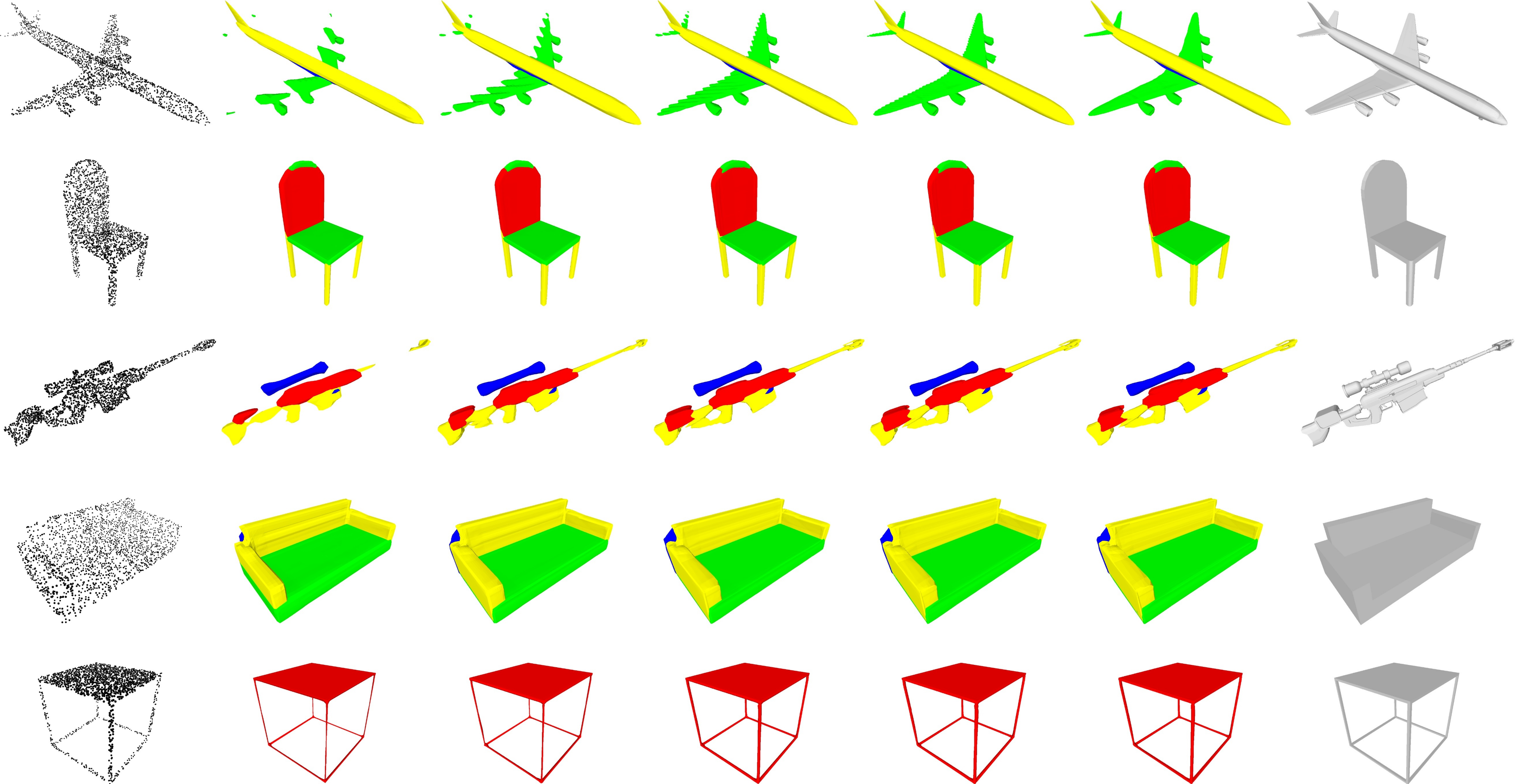}

    \vspace{0.2cm}\hspace{0.1cm}
      \begin{tabular}{p{1.9cm}p{1.9cm}p{1.9cm}p{1.9cm}p{1.9cm}p{1.9cm}p{1.9cm}}
        Input PC &
        \hspace{0.2cm}MC64 &
        \hspace{0.3cm}MC128 &
        \hspace{0.35cm}MC256 &
        \hspace{0.45cm}MC512 &
        \hspace{0.6cm}Ours &
        \hspace{0.8cm}GT \\
      \end{tabular}

    \caption{Qualitative comparisons between analytic marching (AM) and marching cubes (MC) in the context of learning an ensemble of local decoders for reconstructions of novel shape instances. Experiments are conducted on shape instances of five categories from ShapeNet, by training a hypernetwork that gives weights of the ensemble for occupancy modeling; the ensemble is formed by aggregating, via max pooling, outputs of 4 subnetworks, each of which is of depth 4 and width 32. We show an example from each of the five categories. \emph{Results of AM are after simplification of the originally recovered meshes at a ratio of 10\%}. For MC, the sampling resolutions of 3D points range from $64^3$ to $512^3$. Different colors indicate the shape components recovered by individual subnetworks.  }
    \label{FigExpsLocalShapeLearning}

  \end{center}
  \vskip -0.1in
\end{figure*}

We present in Fig. \ref{FigExpsGlobalShapeLearning} and Fig. \ref{FigExpsLocalShapeLearning} the qualitative results of analytic marching respectively by learning a global decoder or an ensemble of local decoders for reconstruction of novel shapes. Quantitative results are respectively presented in Section \ref{ExpGlobalLearning} and Section \ref{ExpLocalLearning} in the main text.

\end{document}